\documentclass{article}

\usepackage[T1]{fontenc}
\usepackage{microtype}
\usepackage{graphicx}
\usepackage{subcaption}
\usepackage{booktabs}
\usepackage{placeins}
\usepackage{array}
\usepackage{multirow}
\usepackage{xcolor}
\usepackage{tikz}
\usetikzlibrary{positioning,arrows.meta,fit,calc,shapes.geometric,backgrounds}
\usepackage{hyperref}

\usepackage[preprint]{icml2026}

\usepackage{amsmath}
\usepackage{amssymb}
\usepackage{mathtools}
\usepackage{amsthm}
\usepackage[capitalize,noabbrev]{cleveref}

\theoremstyle{plain}
\newtheorem{theorem}{Theorem}
\newtheorem{proposition}[theorem]{Proposition}
\newtheorem{lemma}[theorem]{Lemma}

\newtheorem{fact}[theorem]{Fact}
\crefname{fact}{Fact}{Facts}
\Crefname{fact}{Fact}{Facts}
\theoremstyle{definition}

\theoremstyle{remark}

\newcommand{\na}{--}  

\newcommand{\ygg}{Yggdrasil}
\newcommand{\R}{\mathbb{R}}
\newcommand{\Z}{\mathbb{Z}}
\newcommand{\T}{\top}
\newcommand{\h}[1]{h^{(#1)}}          

\newcommand{\diag}{\operatorname{diag}}
\newcommand{\clip}{\operatorname{clip}}
\newcommand{\softmax}{\operatorname{softmax}}
\newcommand{\rms}{\nu}               
\newcommand{\SO}{\mathrm{SO}}
\newcommand{\Ohm}{\mathrm{O}}
\newcommand{\low}{\mathrm{low}}

\newcommand{\out}{\mathrm{out}}

\AtBeginDocument{\raggedbottom}  
\icmltitlerunning{\smash{Yggdrasil}}

\begin{document}

\twocolumn[
  \icmltitle{Yggdrasil: Tree-Routed Recurrent Experts for State Tracking and Fast CPU Inference}

  \begin{icmlauthorlist}
    \icmlauthor{Cosmo Santoni}{imperial}
  \end{icmlauthorlist}

  \icmlaffiliation{imperial}{Imperial College London, London, United Kingdom}
  \icmlcorrespondingauthor{Cosmo Santoni}{cosmo.santoni@ic.ac.uk}

  \icmlkeywords{Sequence Models, Linear Recurrences, State Tracking, Conditional Memory, CPU Inference}

  \vskip 0.12in
]

\printAffiliationsAndNotice{}
\hypersetup{pdftitle={Yggdrasil: Tree-Routed Recurrent Experts for State Tracking and Fast CPU Inference},pdfsubject={Preprint},pdfauthor={Cosmo Santoni}}

\begin{abstract}
Parallel-trainable recurrences such as Mamba, linear attention, and DeltaNet 
offer constant-cost streaming inference, but they suffer from two limitations: 
1) they struggle to track discrete state, and 2) they read every parameter at every 
step, tying model capacity directly to CPU latency. We introduce \ygg{}, a 
recurrent architecture that decouples capacity from per-step compute by looking 
up each layer's transition in a large, tree-routed bank of recurrent experts. 
Each router addresses its bank using the current token and the state of the 
layer below. We prove that this cross-layer context is necessary—a single 
token-indexed layer of non-expanding transitions with finitely many reachable 
states cannot perform conditional resets—while avoiding feedback from the 
layer's own past preserves parallel-scan training. Because each step traverses 
one tree path and reads a single expert per layer, active weights stay within a 
few hundred kilobytes in CPU L2 cache while the full bank resides in RAM. 
Among the parallel-trainable models we evaluate, \ygg{} is the only one to 
solve all five state-tracking tasks. At matched active parameter counts, it 
achieves the lowest test bits on enwik8 (1.540 bpb) and the human genome hg38 
(1.567 bpb), while stepping $3.0\times$ and $3.3\times$ faster on a single CPU 
core (6.7\,$\mu$s and 5.8\,$\mu$s) than the fastest streaming baseline, and delivering the highest GPU throughput of all models at batch 1024. On the 
Genomic Benchmarks, it is the strongest causal model at matched active 
parameters (86.0\% mean accuracy).
\end{abstract}

\section{Introduction}\label{sec:intro}

Many sequence tasks must score symbols one at a time as they arrive, including base-by-base DNA modelling, byte-level compression, and on-device inference \citep{hutter2006enwik8,zhang2025l3tc,gerganov2023llamacpp,gemma3n2025}. Because these workloads cannot batch over time, their cost is governed by a single step at batch size one, ideally running at constant cost per position on one CPU core \citep{williams2009roofline}. Base-level genomic models like HyenaDNA model DNA causally without a tokeniser \citep{nguyen2023hyenadna}, but recompute their context window at every base unless distilled into a recurrence \citep{massaroli2023laughing}. Parallel-trainable recurrences such as Mamba \citep{gu2024mamba,dao2024mamba2}, linear attention \citep{katharopoulos2020transformers,schlag2021linear}, and the DeltaNet family \citep{yang2024deltanet,yang2025gdn,siems2025deltaproduct} do provide constant-cost streaming steps. Yet they hit two fundamental walls. First, they struggle to track discrete state, and none of the parallel-trainable baselines solves all five state-tracking tasks we evaluate ($S_5$, $A_5$, parity, cycle navigation, and modular arithmetic with operator precedence). The strongest, DeltaProduct-4, solves four (\cref{tab:state}), whereas LSTMs and GRUs solve all five at the cost of sequential training \citep{hochreiter1997lstm,cho2014gru,deletang2023chomsky}. Second, they are dense, reading every parameter at every step so that any added capacity directly increases per-token latency.

\begin{figure*}[t]
\centering
\includegraphics[width=0.92\textwidth]{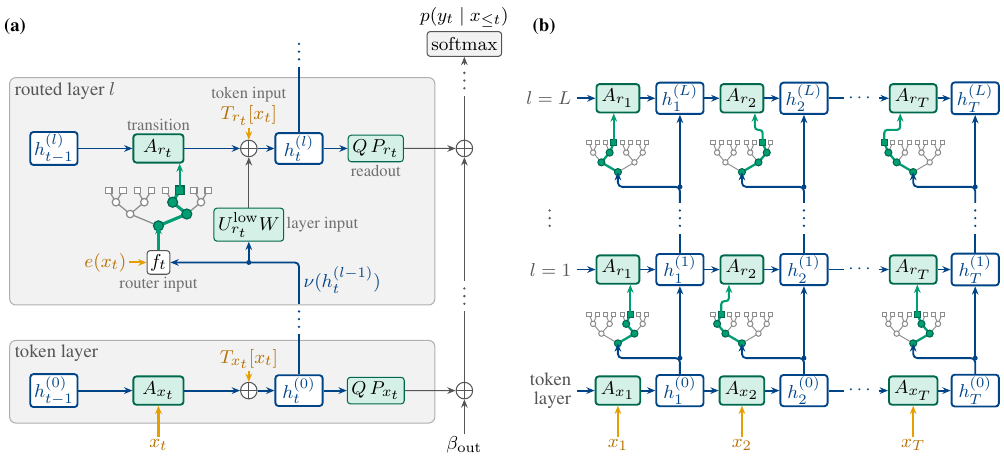}
\caption{\textbf{\ygg{} looks up each layer's transition instead of computing it.} Orange marks the token, green an expert lookup, blue the state and grey the parameters shared within a layer, and layer superscripts are dropped inside each layer. \textbf{(a)}~One inference step. The token layer applies the transition stored for $x_t$. In each routed layer a depth-$D$ tree router reads $f_t=[e(x_t);\nu(h^{(l-1)}_t)]$ and selects expert $r_t$, which supplies the transition $A_{r_t}$, the token input $T_{r_t}[x_t]$, the layer-input map $U^{\mathrm{low}}_{r_t}$ and the readout $P_{r_t}$. The readouts of all layers are summed before the softmax. \textbf{(b)}~The stack unrolled over time, with token inputs and readouts omitted. Each router reads the layer below at the same step and never its own layer's past, so the selected transitions change from step to step while each layer still trains as a parallel scan.}
\label{fig:arch}
\end{figure*}

We introduce \ygg{}, a recurrent architecture that breaks both walls with a single mechanism, looking up each layer's transition in a large bank of recurrent experts rather than computing it (\cref{fig:arch}). Each routed layer stores $2^D$ experts, and a depth-$D$ binary tree selects one in $D$ comparisons to supply the transition, input, and readout for that step. The tree computes its address from the current token and the state of the layer below. Because this address never reads the layer's own past, each layer remains an affine recurrence and trains as an associative parallel scan \citep{blelloch1990prefix,martin2018parallelizing}.

This design directly satisfies the three requirements for exact state tracking. First, transitions must be non-diagonal, since diagonal recurrences cannot track non-solvable groups like $S_5$ and $A_5$ \citep{merrill2024illusion,shakerinava2026diagonal}; \cref{prop:diagonal} establishes the finite-state form of this limit for stacks of diagonal layers. Second, transitions need negative or complex eigenvalues, as eigenvalues restricted to $[0,1]$ cannot track parity \citep{grazzi2025unlocking}. Third, conditional resets require transition selection that depends on more than the current token, and we prove that a single token-indexed layer of non-expanding transitions with finitely many reachable states cannot even retain the first symbol of its input (\cref{thm:contextual-memory}). \ygg{} combines block rotations with signed decays—giving non-diagonal transitions with eigenvalues outside $[0,1]$—while cross-layer routing supplies the context needed for conditional resets. Every ingredient is necessary in trained models. Removing the rotations sends $S_5$ and $A_5$ to chance, and using a shared transition or token-only routing sends modular arithmetic to chance (\cref{sec:exp-ablation}).

The same lookup makes \ygg{} hardware-aware. Like Mamba \citep{gu2024mamba}, it is built around the memory hierarchy, but for CPU inference rather than GPU training (\cref{sec:step-cost}). Because each step reads only one expert per layer, its active weights fit inside the cache of a single CPU core while the full bank resides in RAM. Mixtures of experts and memory layers also decouple storage from compute \citep{shazeer2017moe,fedus2022switch,lample2019pkm,berges2024memorylayers,cheng2026engram}, but only in layers that look up outputs. \ygg{} moves sparse lookup into the recurrence itself; replacing its recurrent experts with output-only experts loses 0.244 bits per byte at equal storage (\cref{sec:exp-ablation}). Prior work on input-switched affine networks, second-order RNNs, and PD-SSM also selects transitions per step \citep{foerster2017isan,giles1992second,omlin1996constructing,terzic2025pdssm}. \ygg{} is the first to look up recurrent transitions in a large tree-routed bank using lower-layer context, preserving both parallel-scan training and a cache-sized CPU step.

\paragraph{Contributions.}
\begin{enumerate}
  \item \textbf{Tree-routed recurrent experts.} \ygg{} looks up each routed layer's transition in a bank of $2^D$ experts, requiring $O(D)$ routing work and cache-sized weight reads per step while preserving parallel-scan training (\cref{sec:model}).
  \item \textbf{State-tracking and sparse-access theory.} We prove that a single token-indexed layer of non-expanding transitions with finitely many reachable states cannot perform conditional resets. We give exact constructions tracking $S_5$ and $A_5$ in 8-bit integer arithmetic and every finite automaton, plus a counting separation between sparse and dense models in weights read per step (\cref{sec:expressivity}).
  \item \textbf{Accuracy.} \ygg{} is the only parallel-trainable model in \cref{tab:state} to solve all five state-tracking tasks. It achieves the lowest test bits on hg38 and enwik8 at every matched active size (\cref{tab:enwik8-full,tab:dna-full}), with its smallest models outperforming every dense baseline at roughly $10\times$ the active parameter count. At 470k active parameters (307.3M stored), it averages 86.0\% across the eight Genomic Benchmarks tasks (five splits each) under the Caduceus protocol \citep{gresova2023genomic,schiff2024caduceus}, making it the best causal model at matched active parameters.
  \item \textbf{Speed.} \ygg{} is the fastest model on one CPU core, with its largest models stepping $3.3\times$ and $3.0\times$ faster than the fastest streaming baseline, and achieves the highest GPU throughput at batch 1024 across all sizes on both datasets (\cref{tab:gpu}). At equal stored parameters, its CPU step is $162\times$--$13{,}566\times$ faster than dense models, and at least $3{,}289\times$ faster at 730M (\cref{sec:experiments}).
\end{enumerate}

\section{Background and related work}\label{sec:background}

\paragraph{Linear recurrences.}
An affine recurrence updates a state $h_t\in\R^n$ as $h_t=\bar A_t h_{t-1}+\bar u_t$. When the coefficients do not depend on $h_{t-1}$, affine composition is associative and all states can be computed in parallel over time \citep{blelloch1990prefix,martin2018parallelizing}. The transition is time-invariant in S4, S4D, S5, and the LRU \citep{gu2022s4,gu2022s4d,smith2023s5,orvieto2023lru}, and input-dependent in the Mamba family \citep{gu2024mamba,dao2024mamba2,lahoti2026mamba3}. Linear attention corresponds to a scalar or identity transition with a rank-one write \citep{katharopoulos2020transformers,schlag2021linear}. To move beyond diagonal transitions, recent work uses delta rules, Householder products, one-hot column transitions, structured linear CDEs, and factored recurrent maps \citep{schlag2021linear,yang2024deltanet,yang2025gdn,siems2025deltaproduct,terzic2025pdssm,walker2025slice,sutskever2011mrnn}, while unitary RNNs constrain the dynamics to preserve norm \citep{arjovsky2016unitary,lezcano2019cheap}. All of these models—like RWKV-7 \citep{peng2025rwkv7}—read every parameter at every step. By contrast, GRUs, LSTMs, and the sLSTM of xLSTM gate on the state itself and must train sequentially, whereas xLSTM's mLSTM gates only on the input and trains in parallel \citep{cho2014gru,hochreiter1997lstm,beck2024xlstm}.

\paragraph{State tracking.}
A model \emph{tracks} a task if the correct label is the unique logit maximiser at every labelled step of every valid input sequence. Its \emph{margin} is the smallest gap between the correct logit and any other logit; in our constructions, this margin is positive and independent of sequence length. Group tasks require maintaining a running product, and they expose the limits of diagonal recurrences, since positive diagonal transitions cannot track parity and general diagonal transitions cannot track non-solvable groups \citep{sarrof2024expressive,merrill2024illusion,grazzi2025unlocking,shakerinava2026diagonal}. Non-diagonal recurrences overcome this barrier \citep{grazzi2025unlocking,siems2025deltaproduct,terzic2025pdssm}. However, group tasks only permute states, whereas general finite automata also reset them. In a permutation--reset automaton, each input either permutes the states or resets all of them to a single state. By Krohn--Rhodes theory, every finite automaton is the image of a cascade of such automata arranged in levels, where each level reads the token and the previous states of lower levels \citep{krohn1965,eilenberg1976,maler1994,liu2023shortcuts}. The routed layers of \ygg{} realise this cascade with one level per layer (\cref{thm:automata}). \Cref{app:related} discusses additional work on state tracking in linear recurrences.

\paragraph{Transition selection.}

\paragraph{Sparse parameter access.}
Mixtures of experts, hash layers, product-key memories, memory layers, and Engram decouple stored capacity from active computation \citep{shazeer2017moe,fedus2022switch,roller2021hash,lample2019pkm,berges2024memorylayers,cheng2026engram}. Token-indexed tables provide another form of sparse access \citep{sadhukhan2026stem,gemma3n2025,zhou2026tngram,tseng2026l3}, while hierarchical experts, soft decision trees, and fast feedforward networks use trees for conditional computation \citep{jordan1994hme,frosst2017soft,belcak2023fff,cho2014exponential}. All of these methods retrieve a value that enters the current step's output; \ygg{} instead retrieves the transition of its recurrence, a difference we compare at equal storage in \cref{sec:exp-ablation}. Following mixture-of-experts convention, we distinguish \emph{stored parameters} (the full model, determining memory footprint) from \emph{active parameters} (those read per token, determining step latency and compute). \ygg{} is built around this split, keeping active parameters in one CPU core's cache while the stored bank resides in RAM.

\section{\ygg{}}\label{sec:model}

\ygg{} stacks one token layer and $L$ routed recurrent layers, each maintaining a state $\h{l}_t\in\R^n$ (\cref{fig:arch}). At every step, each layer applies an \emph{expert} that supplies its transition, additive input, and readout. The token layer selects its expert directly from the token $x_t\in\{1,\dots,V\}$. Each routed layer instead selects one of $2^D$ experts from $x_t$ and the updated state $\h{l-1}_t$ of the layer below.

\subsection{Contextual transition lookup}

Given a selected expert $r=r^{(l)}_t$, layer $l$ updates and reads out its state using only that expert's parameters:
\begin{equation}\label{eq:core}
\h{l}_t=A_r\h{l}_{t-1}+\Gamma_r u^{(l)}_t,
\qquad o^{(l)}_t=P_r\h{l}_t.
\end{equation}
Here $u^{(l)}_t$ packs the token code, a constant coordinate, and a projection of the normalised state from the layer below. Because neither the expert address nor $u^{(l)}_t$ depends on $\h{l}_{t-1}$, the update remains affine in the layer's own state.

\paragraph{Routing.}\label{sec:branches}
Each routed layer's tree router reads the feature vector
\begin{equation}\label{eq:features}
f^{(l)}_t=[e^{(l)}(x_t);\rms(\h{l-1}_t)],
\end{equation}
where $e^{(l)}$ is a learned token embedding and $\rms$ is RMS normalisation. At each internal node $\eta$, the router branches right if $w_\eta^{\T}f^{(l)}_t+\beta_\eta>0$ and left otherwise, reaching one of $2^D$ experts in $D$ comparisons. The token layer has no layer below it and uses $r^{(0)}_t=x_t$, forming an input-switched affine recurrence; a token-only tree router could always be collapsed into this direct lookup (\cref{sec:root}).

\paragraph{Transitions.}\label{sec:leaves}
To support exact state operations while remaining non-expanding, each transition combines coordinatewise decay with block rotations:
\begin{equation}\label{eq:transition}
A_r=\diag(\clip(d_r,-1,1))\,
\big(e^{\Omega_{r,1}}\oplus\cdots\oplus e^{\Omega_{r,G}}\big),
\end{equation}
with $G=n/b$ blocks of size $b$ and skew-symmetric generators $\Omega_{r,g}=\tfrac12(S_{r,g}-S_{r,g}^{\T})$. Because each $e^{\Omega_{r,g}}$ is orthogonal, $\|A_r\|_2\le1$, and \cref{prop:bounded} bounds state growth under bounded additive inputs. Decays in $\{\pm1\}$ realise every orthogonal map within a block, while zero decays add signed partial permutations (\cref{fact:set}), which are the exact operations used in our constructions (\cref{sec:expressivity}).

\paragraph{Inputs and outputs.}\label{sec:output}
The additive input sums a token term $T_r[x_t]=U_rv(x_t)+c_r$ and a lower-layer term $U^{\low}_rW^{(l)}\rms(\h{l-1}_t)$, where the shared matrix $W^{(l)}$ compresses the lower state to rank $k_{\low}$ and the expert matrix $U^{\low}_r$ expands it. For finite vocabularies, $T_r[x_t]$ can be materialised as table rows, though our runtime also supports the factored form (\cref{app:timing}). A shared matrix $Q^{(l)}$ projects each expert readout $o^{(l)}_t\in\R^R$ to the $C$ output classes. Summing all layers' contributions with an output bias before the softmax allows each selected expert to control both its layer's state update and its contribution to the prediction. \Cref{app:architecture} gives the full parameterisation.

\subsection{Parallel evaluation and training}\label{sec:training}

Because layer $l$'s router and additive input depend only on the token and layer $l-1$, they can be evaluated at every position in parallel once layer $l-1$ is computed. Only the affine recurrence in \cref{eq:core} runs over time. Affine composition obeys
\[
(A_2,u_2)\bullet(A_1,u_1)
=(A_2A_1,\,A_2u_1+u_2),
\]
which is associative and yields an $O(\log T)$-depth parallel scan per layer, for a total depth of $O((L+1)\log T)$ across the stack (\cref{prop:scan}). Routing never feeds back into the layer that routes, so once routes are selected, each layer trains like a standard linear recurrence. Moreover, because routing reads $\h{l-1}_t$ after its update with $x_t$, streaming inference and parallel training visit the layers in the exact same order.

\paragraph{Hard routing.}
The forward pass uses a single hard-selected expert from the start of training, perturbed by router noise. A straight-through estimator computes routing gradients from a soft top-$B$ over tree paths at backward temperature $0.5$ \citep{frosst2017soft,bengio2013estimating}. The non-selected experts enter only this gradient and never the forward recurrence (\cref{app:training}).

\paragraph{Hierarchical residual experts.}\label{sec:increments}
To share gradient updates across sparsely selected experts, every node in the router tree (including leaves) stores a residual $\Delta_\eta$ for each expert parameter. Expert $r$ sums the residuals along its root-to-leaf path:
\begin{equation}\label{eq:increments}
\theta_r=\frac{1}{\sqrt{D+1}}\sum_{j=0}^{D}\Delta_{\eta_j(r)},
\end{equation}
where $\eta_j(r)$ is the node at depth $j$. We initialise the root from a single expert and set all deeper residuals to zero. Each routed layer therefore starts as a single shared transition whose experts specialise only where training splits them. Internal nodes learn from all paths passing through them, while deeper residuals allow individual experts to diverge. Because independent leaf residuals can realise any expert bank, this parameterisation preserves the function class (\cref{app:increments}). At inference, we materialise the sums once into the stored expert bank.

\subsection{Hardware-aware design}\label{sec:step-cost}

\ygg{} ties per-step cost to the memory hierarchy rather than total parameter count. For fixed width, block size, input ranks, and output size, a routed layer stores $\Theta(2^D)$ parameters but performs only $O(D)$ work per token, because each step reads $D$ router nodes and one expert, while the transition, lower-layer input, and readout cost $O(nb)$, $O(nk_{\low})$, and $O(nR+CR)$ regardless of bank size (\cref{tab:cost}). Doubling the expert bank adds just one node comparison per routed layer, trading $\Theta(2^D)$ storage for $O(D)$ compute.

On a CPU core, this trade keeps the active working set in cache. Our medium and large models read roughly 100 and 280\,KB of 8-bit weights per token, which fits comfortably inside a single core's 2\,MB L2 cache, while the full bank of $2^D$ experts stays in RAM (\cref{app:timing}). Across all sizes, a step touches between 30 and 281\,KB (under 0.2\% of stored weights), running $2.6\times$ to $5\times$ faster than the same step with cold caches (\cref{app:cache}). On a GPU, the step executes as a single fused kernel over precomputed expert banks, achieving the highest throughput of all evaluated models at batch size 1024 (\cref{app:gpu}).

\section{State tracking and conditional computation}\label{sec:expressivity}

Three mechanisms drive \ygg{}'s expressivity: its transitions realise exact discrete dynamics, its tree routers decide when to apply them, and sparse access lets the model store far more dynamics than any single step reads. We analyse these three properties in turn for the architecture of \cref{sec:model}, deferring proofs and resource bounds to \cref{app:proofs}.

\subsection{Exact permutation tracking}

Let each token specify a permutation matrix $\Pi_x$ on five coordinates, and set $v=(-2,-1,0,1,2)^{\T}$. A token-layer expert with transition $\Pi_x$ and input $\Pi_xv-v$ produces, starting from $h_0=0$,
\[
\begin{aligned}
h_t&=\Pi_{x_t}h_{t-1}+\Pi_{x_t}v-v\\
&=\Pi_{x_t}\cdots\Pi_{x_1}v-v.
\end{aligned}
\]
Because the entries of $v$ are distinct, $h_t$ uniquely identifies the composed permutation. A linear readout then assigns the largest logit to the corresponding group element with a positive margin independent of sequence length (\cref{thm:groups}). This construction tracks both $S_5$ and its subgroup $A_5$ exactly. Moreover, because its states, inputs, materialised transitions, and readout coefficients all fit in 8-bit integers, 32-bit accumulation evaluates the updates and logits without error at any length.

The same construction extends to any finite group with a faithful orthogonal representation of dimension $m\le b$, tracked by a single block with readout rank $R\ge m$, which covers parity and cycle navigation as special cases. Parity requires the eigenvalue $-1$, which an expert achieves via either a negative decay or a $\pi$-rotation of a two-dimensional block ($-I$). Non-solvable groups, however, require off-diagonal transitions, because no stack of diagonal affine layers with finitely many reachable states per layer can track $A_5$, even if each layer selects its coefficients from the token and all states below it (\cref{prop:diagonal}). This establishes a finite-state counterpart to bounded-precision limits on recurrent models \citep{merrill2024illusion,grazzi2025unlocking,shakerinava2026diagonal}.

\subsection{Contextual writes and finite automata}\label{sec:resets}

A token-indexed affine map can reset its state on a designated symbol by setting $A_x=0$. Writing the first symbol and preserving it across all later steps is harder, because the same symbol must trigger a write at $t=1$ and leave the state untouched afterwards. A single token layer with non-expanding transitions and finitely many reachable states cannot do this, whereas adding one routed layer can.

\begin{theorem}[Contextual memory]\label{thm:contextual-memory}
A single token-only affine layer whose transitions are non-expanding in the Euclidean norm, $\|A_x\|_2\le1$, and which reaches finitely many states cannot track the first symbol of an arbitrary binary input, even with an arbitrary readout depending on $(h_t,x_t)$. A token layer and one routed layer with block size $b\ge2$ and router depth $D=2$ can track both the first-symbol task and even pairs, whose label is $[x_t\ne x_1]$.
\end{theorem}

The positive construction uses two token-layer coordinates to separate the first step from all later steps. Reading this context, the routed layer writes the current symbol at $t=1$ and applies the identity thereafter, without ever depending on its own past state. The impossibility result is a special case of \cref{thm:noreset}, and the construction appears in \cref{prop:evenpairs}. The same obstruction governs modular arithmetic with operator precedence, which a token layer and one routed layer of depth $7$ track modulo $5$.

Permutations and contextual resets suffice for any finite automaton. In its Krohn--Rhodes cascade (\cref{sec:background}), every component either permutes its states or resets to a fixed state, selected by the token and the states of lower levels.

\begin{theorem}[Finite automata]\label{thm:automata}
Every finite automaton can be tracked by \ygg{} with sufficiently large state width, block size, readout rank and router depth. If the automaton admits a permutation--reset cascade with $\ell$ levels, $\ell+1$ routed layers suffice.
\end{theorem}

The proof assigns one routed layer to each cascade level and implements its permutations and resets as partial permutations, with bank size $2^D\ge2V\prod_i m_i$ for components of sizes $m_i$ (\cref{app:proof-automata}).

\subsection{A memory--computation separation}

Sparse access lets a model query a large transition table without reading all of it at every step. We formalise this advantage using a family of hidden-counter tasks. Each token carries an action $a_t\in\{1,\dots,K\}$ and a counter increment, and the label is a running product in a fixed finite group $\mathcal G$ whose factor is looked up in a table $\varphi$ indexed by the action and the counter modulo $M$, yielding $|\mathcal G|^{KM}$ distinct tasks (\cref{lem:distinct}).

When $M$ is the product of the first $I$ primes, a token layer tracks the counter via its residues and a single routed layer retrieves the corresponding table entry. State width, block size, router depth, and per-token work all grow only polynomially in $\log M$ and $\log K$ (\cref{thm:separation}). By contrast, any dense architecture with $\omega$ bits per weight that reads all of its weights at every step requires
\begin{equation}\label{eq:dense-lower-bound}
\text{work per token}\ \ge\
\frac{M\log_2|\mathcal G|-c/K}{2\omega}
\end{equation}
to realise a fraction $2^{-c}$ of the family, which grows linearly in $M$. \ygg{} pays for a larger task family once in stored experts, whereas a dense model pays at every step. The separation thus lies in weights read per step, not weights stored (\cref{app:proof-separation}).

\begin{table*}[t]
\centering
\caption{\textbf{State tracking.} Final-answer accuracy on test sequences of lengths 40 to 256 after $100\mathrm{k}$ training steps on lengths 3 to 40. Chance is 0.008 on $S_5$, 0.017 on $A_5$, 0.5 on parity and 0.2 on cycle navigation and modular arithmetic. Best of 3 seeds by validation. The search selected the size of every model per task (\cref{app:baselines}). DeltaProduct uses 2 or 4 Householder factors per step and negative eigenvalues. Sequential models train step by step and parallel models over the whole sequence at once. Bold marks a solved task (raw accuracy $\ge0.90$, \citet{deletang2023chomsky}).}
\label{tab:state}
\footnotesize
\setlength{\tabcolsep}{3pt}
\begin{tabular}{@{}llccccc@{}}
\toprule
& & \multicolumn{2}{c}{\textbf{Non-solvable}} & \multicolumn{2}{c}{\textbf{Abelian}} & \textbf{Reset} \\
\cmidrule(lr){3-4}\cmidrule(lr){5-6}\cmidrule(lr){7-7}
& \textbf{Model} & $\boldsymbol{S_5}$ & $\boldsymbol{A_5}$ & \textbf{Parity} & \textbf{Cycle} & \textbf{Mod.\ arith.} \\
&  & {\tiny\citep{merrill2024illusion}} & {\tiny\citep{merrill2024illusion}} & {\tiny\citep{deletang2023chomsky}} & {\tiny\citep{deletang2023chomsky}} & {\tiny\citep{deletang2023chomsky}} \\
\midrule
\multirow{2}{*}{Sequential} & GRU & \textbf{0.979} & \textbf{1.000} & \textbf{1.000} & \textbf{1.000} & \textbf{1.000} \\
 & LSTM & \textbf{0.995} & \textbf{0.999} & \textbf{1.000} & \textbf{1.000} & \textbf{1.000} \\
\cmidrule(lr){1-7}
\multirow{7}{*}{Parallel} & Transformer & 0.010 & 0.031 & 0.506 & 0.238 & 0.225 \\
 & Mamba-2 & 0.010 & 0.017 & 0.569 & 0.650 & 0.395 \\
 & Gated DeltaNet & 0.018 & \textbf{0.999} & \textbf{0.999} & 0.615 & 0.826 \\
 & DeltaProduct-2 & 0.688 & \textbf{0.999} & \textbf{1.000} & 0.713 & \textbf{1.000} \\
 & DeltaProduct-4 & \textbf{0.935} & \textbf{1.000} & \textbf{1.000} & 0.684 & \textbf{1.000} \\
 & PD-SSM & 0.009 & 0.020 & 0.807 & \textbf{0.986} & \textbf{1.000} \\
 & \ygg{} & \textbf{1.000} & \textbf{0.980} & \textbf{1.000} & \textbf{1.000} & \textbf{0.981} \\
\bottomrule
\end{tabular}
\end{table*}

\section{Experiments}\label{sec:experiments}

We evaluate \ygg{} across four axes: length generalisation on state tracking, accuracy and latency on DNA and text at matched active parameter counts, transfer to DNA classification on the Genomic Benchmarks, and scaling plus ablations.

\subsection{Protocol}

\textbf{Tasks.} Our state-tracking suite covers the word problems of $S_5$ and $A_5$ \citep{merrill2024illusion,grazzi2025unlocking,siems2025deltaproduct}, parity, cycle navigation, and modular arithmetic with operator precedence \citep{deletang2023chomsky}, trained on lengths 3--40 and tested on lengths 40--256. Next-token prediction uses base-level hg38 and byte-level enwik8 (\cref{app:setup}).

\textbf{Models and sizes.} We compare against official public implementations of GRU, LSTM, Mamba-2, Mamba-2 + PKM, a Llama-style Transformer, Gated DeltaNet, DeltaProduct (with 2 or 4 Householder factors), PD-SSM, and HyenaDNA \citep{cho2014gru,hochreiter1997lstm,dao2024mamba2,lample2019pkm,touvron2023llama,vaswani2017attention,yang2025gdn,siems2025deltaproduct,terzic2025pdssm,nguyen2023hyenadna} (\cref{app:baselines}). Text models target 35k, 105k, and 330k active parameters; DNA models target 27k, 90k, and 290k. Every model sits within 10\% of its target using its native architecture, with \ygg{} accessing at most 0.22\% of its stored parameters per token. On state tracking, hyperparameter search also selects model size.

\textbf{Training and selection.} For every model and task, we run eight Bayesian-search trials on one seed, selecting by validation bits on text and DNA and by out-of-distribution validation accuracy on state tracking. We then train the chosen configuration across three seeds and report the seed with the best validation score (\cref{app:protocol}). We follow the length split of \citet{walker2025slice} but report accuracy on held-out test sequences rather than the best validation accuracy reached during training.

\textbf{Metrics.} We score text and DNA by test bits per token of the float model over the full test split. State tracking is scored by final-answer accuracy averaged over test lengths, with $\ge 0.90$ marking a solved task \citep{deletang2023chomsky}.

\textbf{Latency.} All CPU timings measure one batch-1 step computing the full output distribution on a single pinned core. \ygg{} runs in our 8-bit integer kernel \citep{jacob2018quantization}, and each baseline runs in its fastest public runtime (including 8-bit quantised runtimes) that matches its float accuracy (\cref{app:timing}).

\subsection{State tracking}\label{sec:exp-state}

\ygg{} is the only parallel-trainable model in \cref{tab:state} to solve all five state-tracking tasks. Among the other parallel baselines, DeltaProduct-4 solves four, DeltaProduct-2 solves three, Gated DeltaNet and PD-SSM solve two, and Transformer and Mamba-2 solve none. Only PD-SSM and \ygg{} among the parallel models solve cycle navigation. GRU and LSTM also solve all five tasks, but require sequential training.

\subsection{DNA and text}\label{sec:exp-lm}

\begin{table}[t]
\centering
\caption{\textbf{Genomic Benchmarks, causal models.} Test accuracy (\%), mean over 5 splits with the range over the splits after $\pm$, at about 470k parameters (active for \ygg{}, which stores 307.3M total parameters). HyenaDNA, Mamba and \ygg{} are pretrained on the human genome and fine-tuned per task. Bold marks the best per row.}
\label{tab:gb}
\footnotesize
\setlength{\tabcolsep}{1pt}
\begin{tabular}{@{}lcccc@{}}
\toprule
 & \multicolumn{3}{c}{\textbf{Published}} & \\
\cmidrule(lr){2-4}
\textbf{Task} & \textbf{CNN} & \textbf{HyenaDNA} & \textbf{Mamba} & \textbf{\ygg{}} \\
{\tiny\citep{gresova2023genomic}} & {\tiny\citep{schiff2024caduceus}} & {\tiny\citep{schiff2024caduceus}} & {\tiny\citep{schiff2024caduceus}} & \\
\midrule
Mouse & 71.5{\tiny$\pm$8.7} & \textbf{78.0}{\tiny$\pm$2.5} & 74.3{\tiny$\pm$5.4} & 74.0{\tiny$\pm$4.1} \\
Coding & 89.2{\tiny$\pm$0.8} & 90.4{\tiny$\pm$0.5} & 90.4{\tiny$\pm$0.4} & \textbf{93.4}{\tiny$\pm$0.6} \\
Hum./worm & 94.2{\tiny$\pm$0.2} & 96.4{\tiny$\pm$0.2} & \textbf{96.7}{\tiny$\pm$0.2} & 96.6{\tiny$\pm$0.2} \\
Cohn & 70.2{\tiny$\pm$2.1} & 72.9{\tiny$\pm$1.4} & \textbf{73.2}{\tiny$\pm$2.9} & 72.9{\tiny$\pm$1.6} \\
Ensembl & 74.4{\tiny$\pm$12.2} & 84.9{\tiny$\pm$0.6} & 86.2{\tiny$\pm$0.8} & \textbf{88.2}{\tiny$\pm$0.6} \\
Reg. & \textbf{87.2}{\tiny$\pm$0.5} & 86.9{\tiny$\pm$1.2} & 81.4{\tiny$\pm$21.1} & 87.1{\tiny$\pm$1.3} \\
OCR & 69.8{\tiny$\pm$1.3} & 78.3{\tiny$\pm$0.7} & \textbf{81.5}{\tiny$\pm$0.2} & 79.6{\tiny$\pm$0.5} \\
Non-TATA & 86.1{\tiny$\pm$0.9} & 94.4{\tiny$\pm$0.2} & 93.3{\tiny$\pm$0.7} & \textbf{96.0}{\tiny$\pm$1.0} \\
\midrule
Mean & 80.3 & 85.3 & 84.6 & \textbf{86.0} \\
\bottomrule
\end{tabular}
\end{table}

\begin{figure*}[t]
\centering
\includegraphics[width=0.76\textwidth]{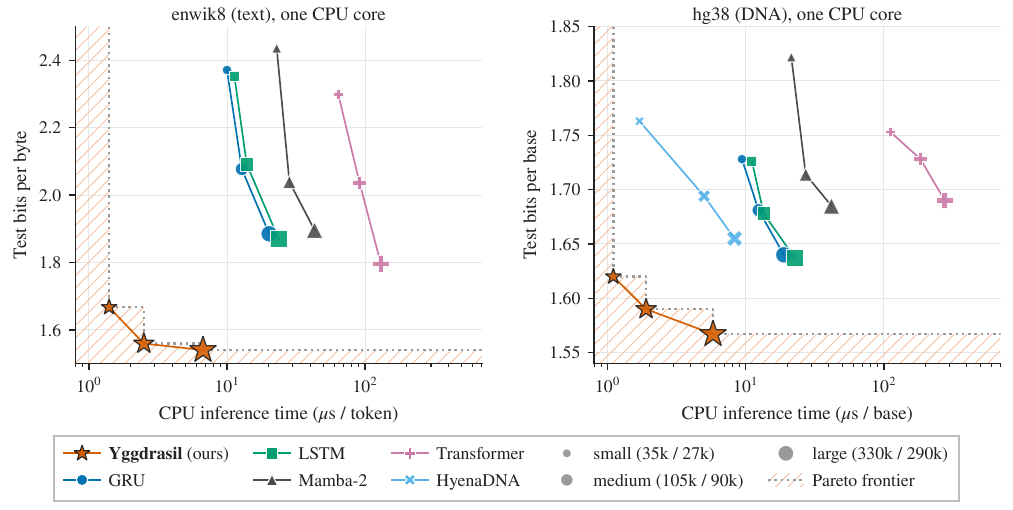}
\caption{\textbf{Accuracy against inference time on one CPU core.} Float test bits per byte on enwik8 (left) and per base on hg38 (right) against microseconds per token. Each line joins one model at the three sizes of active parameters (\cref{tab:enwik8-full,tab:dna-full}), and marker size grows with the size, labelled in the key by its active parameters on enwik8 / hg38. Points are the best of 3 seeds by validation. HyenaDNA is shown at its offline time per base. Gated DeltaNet and DeltaProduct have no public CPU runtime and appear in \cref{fig:pareto-gpu}. The dotted staircase is the Pareto frontier, and no model reaches the hatched region.}
\label{fig:pareto}
\end{figure*}

At matched active parameter counts, \ygg{} achieves the lowest test bits and the lowest CPU latency at all three sizes on both datasets (\cref{fig:pareto,tab:enwik8-full,tab:dna-full}). On hg38, its largest model reaches 1.567 bits per base versus 1.598 for Mamba-2 with a product-key memory (PKM) \citep{lample2019pkm}, the strongest baseline. Its 8-bit kernel steps in 5.8\,$\mu$s, $3.3\times$ faster than GRU at 18.9\,$\mu$s, the fastest streaming baseline. Even HyenaDNA's offline scoring takes $1.4\times$--$2.7\times$ longer per base than \ygg{}'s streaming step across sizes, while recomputing its window per base is orders of magnitude slower (\cref{app:dna}). On enwik8, \ygg{}'s largest model reaches 1.540 bits per byte versus 1.673 for Mamba-2 + PKM, stepping in 6.7\,$\mu$s versus 20.2\,$\mu$s for GRU ($3.0\times$ faster).

The efficiency gap is even starker across scales. With 35.5k active parameters on enwik8 and 25.2k on hg38, our smallest \ygg{} models reach 1.667 and 1.620 bits and outperform every dense baseline at the largest size, despite using $8.5\times$--$9.9\times$ fewer active parameters on enwik8 and $10.8\times$--$11.9\times$ fewer on hg38. Mamba-2 + PKM, matched to \ygg{} in both stored and active parameters, trails at every size on both datasets. Over the full test stream, the timed 8-bit kernel stays within 0.010 bits per token of the float model (\cref{app:timing}). On a GPU, Mamba-2 is faster at batch size 1, whereas at batch size 1024 \ygg{} achieves the highest throughput of all models at every size, running $1.5\times$--$3.0\times$ faster than Mamba-2 (\cref{tab:gpu,fig:pareto-gpu}).

\subsection{Genomic Benchmarks}\label{sec:exp-gb}

We evaluate transfer to DNA classification on the eight tasks of the Genomic Benchmarks \citep{gresova2023genomic} under the Caduceus protocol \citep{schiff2024caduceus}: mouse enhancers (Mouse), coding versus intergenomic sequences (Coding), human versus worm (Hum./worm), Cohn and Ensembl human enhancers (Cohn, Ensembl), regulatory elements (Reg.), open chromatin regions (OCR), and non-TATA promoters (Non-TATA). Because each task classifies a short sequence, a causal streaming model can score it as it arrives. Pretrained on hg38 and fine-tuned per task at roughly 470k active parameters (307.3M stored; \cref{app:gb}), \ygg{} averages 86.0\%, outperforming both HyenaDNA (85.3\%) and Mamba (84.6\%) as the strongest causal model at matched active parameters (\cref{tab:gb}).

\subsection{Scaling and ablations}

\textbf{Speed at equal stored parameters.} Matching models from roughly 5M to 730M stored parameters, \ygg{} is $162\times$--$1{,}740\times$ faster per CPU step than dense baselines at 5M, widening to $3{,}289\times$--$13{,}566\times$ at 730M (\cref{tab:capacity_speed}). Dense models must read every weight on every token, whereas \ygg{} reads only 56k to 172k parameters per step (\cref{app:eqp}).

\textbf{Ablations.} Removing block rotations sends $S_5$ and $A_5$ to chance, and either sharing transitions or routing on the token alone sends modular arithmetic to chance. On enwik8, removing the routed-layer recurrence to store outputs instead costs 0.244 bits per byte at equal storage and token-only routing costs 0.153, whereas sharing a single transition per routed layer stays within 0.003 bits per byte while losing conditional resets on modular arithmetic (\cref{sec:exp-ablation}).

\section{Conclusion}\label{sec:conclusion}

\ygg{} replaces computed recurrent transitions with contextual lookups into a tree-routed bank of experts. Addressing each bank from the token and the state of the layer below yields contextual transitions while preserving parallel-scan training. The resulting architecture tracks permutations, conditional memory, and arbitrary finite automata; solves all five evaluated state-tracking benchmarks in parallel; and keeps active weights in CPU L2 cache while the expert bank sits in RAM, improving both accuracy and CPU latency at matched active parameter counts.

\paragraph{Limitations.}
\ygg{} shifts capacity from per-token compute into stored parameters. Its largest models occupy 681.3\,MB on enwik8 and 146.5\,MB on hg38, which fits easily in laptop RAM but is much larger than the 1.5\,MB footprint of the matched dense baselines. Our main comparisons therefore hold active parameters fixed. Dense models with comparable storage, such as a 41M-parameter Transformer-XL (1.06 bpc on enwik8; \citealp{dai2019transformerxl}), achieve lower bits at the cost of reading every parameter on every step. Finally, our experiments cover up to 470k active parameters and contexts of 256 bytes or 1024 bases; scaling to larger active sizes, longer contexts, subword vocabularies, and attention hybrids remains future work.

\section*{Acknowledgements}
This work was supported by the Wellcome Trust (226072/Z/22/Z) and by
the Medical Research Council (MRC) Centre for Global Infectious Disease
Analysis (reference MR/X020258/1), funded by the United Kingdom MRC.
I thank Akshay Kumar at Radix Trading for generously dedicating his spare time
and effort to this work through discussions that helped shape its
conception, for valuable feedback and pointers, and for suggesting related work. I thank the Research
Software for Infectious Disease Epidemiology (RESIDE) team at Imperial
College London for access to their computing cluster. The GPU compute
was funded by MRC Mid-Range Equipment Funding (UKRI2681). For the
purpose of open access, the author has applied a CC BY public copyright
licence to this submission.

\section*{Software and Data}
{\raggedright Code is available at \url{https://github.com/CosmoNaught/yggdrasil}.\par}
\bibliography{refs}
\bibliographystyle{icml2026}
\newpage
\appendix

\onecolumn

\section{Further related work}\label{app:related}

The linear RNN of \citet{fan2024advancing} computes a block-diagonal transition from each input and extrapolates on regular languages such as even pairs and modular arithmetic, without a bank of stored experts or a tree that addresses one. Linear transformers fail parity, which their recurrent and self-referential extensions solve \citep{irie2023practical}. Depth recovers part of what diagonal transitions lose, since chains of diagonal selective layers approach the expressivity of dense ones as the chain grows \citep{cirone2024theoretical}, and state-space models with non-negative gates recognise exactly the star-free languages at finite precision \citep{sarrof2024expressive}. \citet{grazzi2025unlocking} show that linear recurrences whose transitions are products of generalised Householder matrices, each with eigenvalues in $[-1,1]$, can recognise any regular language, and \ygg{} reaches every finite automaton with one selected transition per layer, drawn from a routed bank and addressed by the layer below.

\section{Proofs}
\label{app:proofs}

\paragraph{Conventions.}
For a word $w=x_1\cdots x_s$ the token layer applies $F_w=F_{x_s}\circ\dots\circ F_{x_1}$ with $F_x(h)=A_xh+T_x[x]$, so $F_{uw}=F_w\circ F_u$, the linear part of $F_w$ is $A_{x_s}\cdots A_{x_1}$, and the state after a prefix $u$ is $h(u)=F_u(0)$.

\emph{Offset coding.} A state $u_t=A_tu_{t-1}+c_t$ with $u_0=v\ne0$ is run as $h_t=u_t-v$, with $h_0=0$ and the constant expert input $A_tv+c_t-v$. The readout $Pu_t=Ph_t+Pv$ puts $Pv$ into $\beta_{\out}$ if all experts share it, and otherwise into the column of $P_r$ that reads a constant coordinate $\kappa$ with transition row $0$, row $\kappa$ of $c_r$ equal to $1$ and row $\kappa$ of $U_r$ and $U^{\low}_r$ equal to $0$, so $h_\kappa=1$ for $t\ge1$.

\emph{Unused parts.} Unused coordinates have transition row $0$ and input $0$, every construction has $U^{\low}_r=0$, and layers other than the one named have output map $0$.

\emph{Token tests and padding.} With $e^{(l)}(x)$ a $\pm1$ binary code of $x$, a node with weight $1$ on one entry and bias $0$ tests that bit, so $\lceil\log_2V\rceil$ levels determine $x_t$. Nodes with $w_\eta=0$ and $\beta_\eta=0$ always go left and pad a tree router to depth $D$, and experts on paths that decode no token are arbitrary.

\subsection{Contraction}
\label{app:proof-norm}

\begin{fact}[Contraction subclass]
\label{fact:norm}
For every expert transition $A_r=\diag(\delta_r)\,(\Theta_{r,1}\oplus\cdots\oplus\Theta_{r,G})$ with rotations $\Theta_{r,g}$ and $|\delta_{r,i}|\le1$, $\|A_r\|_2\le1$, and $\|\sum_r\pi_rA_r\|_2\le1$ for all $\pi_r\ge0$ with $\sum_r\pi_r=1$.
\end{fact}

\begin{proof}[Proof of \cref{fact:norm}]
For skew-symmetric $\Omega$, $\exp(\Omega)^{\T}\exp(\Omega)=\exp(-\Omega)\exp(\Omega)=I$, so every $\Theta_{r,g}$ is orthogonal and the block-diagonal factor has norm $1$. Since $\|\diag(\delta_r)\|_2=\max_i|\delta_{r,i}|\le1$, submultiplicativity gives $\|A_r\|_2\le1$, and the triangle inequality gives $\|\sum_r\pi_rA_r\|_2\le\sum_r\pi_r\|A_r\|_2\le1$.
\end{proof}

\subsection{Transition set}
\label{app:proof-set}

\begin{fact}[Rotation transition set]
\label{fact:set}
The matrices $A_r$ that an expert can express form exactly the set $\{\diag(\delta)\,(\Theta_1\oplus\cdots\oplus\Theta_G):\Theta_g\in\SO(b),\ |\delta_i|\le1\}$. With $\delta\in\{\pm1\}^n$ it contains every block-diagonal orthogonal matrix, and with $\delta\in\{0,\pm1\}^n$ it contains every signed partial permutation matrix whose nonzero entries lie inside the blocks.
\end{fact}

\begin{proof}[Proof of \cref{fact:set}]
Every transition lies in the stated set, since $\clip(d_r,-1,1)\in[-1,1]^n$ and $\exp(\Omega)$ is orthogonal (\cref{app:proof-norm}) with $\det\exp(\Omega)=e^{\operatorname{tr}\Omega}=1$.

Conversely, given $\delta$ and $\Theta_g$, take $d_r=\delta$, which the clip leaves unchanged. By the real normal form, $\Theta\in\SO(b)$ is $X\diag(\mathrm{Rot}_{\phi_1},\dots,\mathrm{Rot}_{\phi_j},\pm1,\dots,\pm1)X^{\T}$ with $X$ orthogonal, and since $\det\Theta=1$ the entries $-1$ come in pairs, each a rotation by $\pi$. As $\mathrm{Rot}_\phi=\exp\begin{psmallmatrix}0&-\phi\\\phi&0\end{psmallmatrix}$, $\Theta=\exp(\Omega)$ with $\Omega=X\Omega'X^{\T}$ skew-symmetric for a block-diagonal skew-symmetric $\Omega'$, and $S_{r,g}=\Omega$ gives $\tfrac12(S_{r,g}-S_{r,g}^{\T})=\Omega$.

\emph{Orthogonal blocks.} If $\Theta\in\Ohm(b)$ has $\det\Theta=-1$, then $\Theta=E\,(E\Theta)$ with $E=\diag(-1,1,\dots,1)$ and $E\Theta\in\SO(b)$, which for each such block gives $\diag(\delta)\,(\Theta'_1\oplus\cdots\oplus\Theta'_G)$ with $\delta\in\{\pm1\}^n$ and $\Theta'_g\in\SO(b)$.

\emph{Signed partial permutations.} Let a $b{\times}b$ block $P$ have at most one nonzero entry, $\pm1$, in each row and column, with nonzero rows $\mathcal I$ and $P_{i,\pi(i)}=\sigma_i$. The injective $\pi$ on $\mathcal I$ extends to a permutation with matrix $\Pi$, and $P=\diag(\varepsilon)\Pi$ with $\varepsilon_i=\sigma_i$ on $\mathcal I$ and $0$ otherwise. By the previous step $\Pi=\diag(\delta')\Theta'$ with $\delta'\in\{\pm1\}^b$ and $\Theta'\in\SO(b)$, so $P=\diag(\varepsilon\odot\delta')\Theta'$ with $\varepsilon\odot\delta'\in\{0,\pm1\}^b$. Applying this to every block proves the claim.
\end{proof}

\subsection{Parallel training}
\label{app:proof-scan}

\begin{proposition}[Parallel training]\label{prop:scan}
Fix the hard or soft expert probabilities of every layer and step. All states of the $L+1$ layers are computed exactly by $L+1$ associative scans, each of depth $O(\log T)$.
\end{proposition}

\begin{proof}[Proof of \cref{prop:scan}]
With the probabilities $\pi_{t,r}$ fixed, layer $l$ follows $\h{l}_t=\bar A_t\h{l}_{t-1}+\bar u_t$ with $\bar A_t=\sum_r\pi_{t,r}A_r$ and $\bar u_t=\sum_r\pi_{t,r}\big(T_r[x_t]+U^{\low}_rW^{(l)}\rms(\h{l-1}_t)\big)$, where the last term is absent in the token layer. Step $t$ is the affine map $(\bar A_t,\bar u_t)$, and $\bullet$ is associative as a composition of maps, so an associative scan of depth $O(\log T)$ with $O(T)$ products gives all prefix products and hence all $\h{l}_t$ from $\h{l}_0=0$ \citep{blelloch1990prefix}. The inputs and, in training, the probabilities of layer $l$ depend only on the tokens and $\h{l-1}_t$, so scanning layers $0,\dots,L$ in order gives sequential depth $O((L+1)\log T)$ \citep{martin2018parallelizing}.
\end{proof}

\subsection{State bound}
\label{app:proof-bounded}

\begin{proposition}[State bound]
\label{prop:bounded}
For its total additive inputs $u_s$, both hard and soft updates of a layer satisfy $\|h_t\|_2\le\|h_0\|_2+\sum_{s\le t}\|u_s\|_2$. If every transition has norm at most $\bar\delta<1$, then $\|h_t\|_2\le\bar\delta^t\|h_0\|_2+\max_s\|u_s\|_2/(1-\bar\delta)$.
\end{proposition}

\begin{proof}[Proof of \cref{prop:bounded}]
By \cref{fact:norm}, $\|\bar A_t\|_2\le1$ for hard transitions and convex mixtures, so $\|h_t\|_2\le\|h_{t-1}\|_2+\|u_t\|_2$, and induction gives the first claim. Under the strict bound, $\|h_t\|_2\le\bar\delta^t\|h_0\|_2+\sum_{s\le t}\bar\delta^{t-s}\|u_s\|_2$, and the geometric series gives the second.
\end{proof}

In a routed layer $\|\rms(h)\|_2<\sqrt n$, so with the finite token table the additive input is bounded uniformly over input sequences.

\subsection{Token-only tree routers in the token layer}
\label{app:proof-roottree}

\begin{proposition}[The token layer needs no tree router]
\label{prop:roottree}
A tree router that reads only the token sends each token to one expert. The token layer, with the expert of token $x$ set to the expert that the tree router selects for $x$, gives the same states and outputs.
\end{proposition}

\begin{proof}[Proof of \cref{prop:roottree}]
A tree router that reads only $e(x_t)$ selects an expert $\lambda(x_t)$. Set $A_x=A_{\lambda(x)}$, $U_x=U_{\lambda(x)}$, $c_x=c_{\lambda(x)}$ and $P^{(0)}_x=P_{\lambda(x)}$. Both layers then apply the same maps at every step.
\end{proof}

\subsection{Routing after normalisation}
\label{app:proof-kappa}

\begin{lemma}[Tests survive the division]
\label{lem:kappa}
Let the previous layer have a coordinate $\kappa$ whose transition row is $0$ and whose total input is $1$ in every expert and for every token (row $\kappa$ of $c_r$ is $1$, and row $\kappa$ of $U_r$ and, in a routed layer, of $U^{\low}_r$ is $0$), so $h_\kappa=1$ after every step. Then for all $w\in\R^n$ and $\beta\in\R$, a node with bias $0$, weight $0$ on the token embedding and weights $w+\beta \mathbf e_\kappa$ on $\rms(h)$ has score $w^{\T}\rms(h)+\beta\,\rms(h)_\kappa=\varsigma\,(w^{\T}h+\beta)$ with $\varsigma>0$, so it routes exactly as the affine test $w^{\T}h+\beta>0$.
\end{lemma}

\begin{proof}[Proof of \cref{lem:kappa}]
Row $\kappa$ of every transition is $0$ and row $\kappa$ of the total input $T_r[x]+U^{\low}_rW\rms(\cdot)$ is $1$, so $h_\kappa=1$. As $\rms(h)=\varsigma\,h$ with $\varsigma=(\tfrac1n\|h\|^2+\epsilon)^{-1/2}>0$, $\rms(h)_\kappa=\varsigma$ and the score $(w+\beta \mathbf e_\kappa)^{\T}\rms(h)$ equals $\varsigma\,(w^{\T}h+\beta)$ for every $w$, including $w_\kappa\ne0$.
\end{proof}

Since $h_\kappa=1$, $\varsigma\le\sqrt n$, and $\|h\|\le H$, a uniform bound because the construction reaches finitely many states, gives $\varsigma\ge(H^2/n+\epsilon)^{-1/2}$, so a test with margin $\mu$ keeps margin at least $\mu(H^2/n+\epsilon)^{-1/2}$. A test with $\beta=0$ needs no $\kappa$.

\subsection{Group tasks}
\label{app:proof-groups}

\begin{theorem}[Groups]
\label{thm:groups}
Let $\mathcal G$ be a finite group, let token $x$ stand for $g_x\in\mathcal G$, and let $\rho:\mathcal G\to\Ohm(m)$ be a faithful orthogonal representation with $m\le b$. Then the token layer, using one block and readout rank $R\ge m$, tracks the running product $g_t=g_{x_t}\cdots g_{x_1}$.
\end{theorem}

\begin{proof}[Proof of \cref{thm:groups}]
\emph{A vector with trivial stabiliser.} For $g\ne\mathrm{id}$, $\rho(g)\ne I$ by faithfulness, so the fixed space of $\rho(g)$ is a proper subspace, and as a finite union of these has measure zero, some $v$ has $\rho(g)v\ne v$ for all $g\ne\mathrm{id}$. Then $\rho(g)v=\rho(g')v$ implies $g=g'$, so the $\rho(g)v$ are distinct, all of length $\|v\|$.

\emph{Experts and state.} On the first $m$ coordinates of block $1$, expert $x$ has transition $\rho(g_x)\oplus I_{b-m}$, an expert transition by \cref{fact:set}, and input $T_x[x]=\rho(g_x)v-v$. This is the offset coding of $u_t=\rho(g_t)v$ with $u_0=v$, since $h_0=0=\rho(\mathrm{id})v-v$ and, inductively,
$h_t=\rho(g_{x_t})\big(\rho(g_{t-1})v-v\big)+\rho(g_{x_t})v-v=\rho(g_t)v-v$.

\emph{Output.} Take $P^{(0)}_x=\begin{psmallmatrix}I_m&0\\0&0\end{psmallmatrix}\in\R^{R\times n}$ for every $x$, possible as $R\ge m$, row $g$ of $Q^{(0)}$ equal to $\big((\rho(g)v)^{\T},0\big)$ and $\beta_{\out,g}=\langle\rho(g)v,v\rangle$. The logit of class $g$ is $\langle\rho(g)v,\rho(g_t)v\rangle$, and $\|v\|^2-\langle\rho(g)v,\rho(g_t)v\rangle=\tfrac12\|\rho(g)v-\rho(g_t)v\|^2$ is $0$ for $g=g_t$ and at least $\mu=\tfrac12\min_{g\ne\mathrm{id}}\|\rho(g)v-v\|^2>0$ otherwise, a margin $\mu$ at every length.
\end{proof}

\emph{Task weights.} Parity is $\Z_2$ with $\rho(1)=-1$ on $\R^1$ and $v=1$. Cycle navigation moves by $-1$, $0$ or $+1$ in $\Z_5$, with $\rho$ the $5{\times}5$ cyclic permutation matrices (or the rotations by multiples of $2\pi/5$ in a plane). For $S_5$ and $A_5$, $\rho$ is the $5{\times}5$ permutation matrices, for which any $v$ with distinct entries has trivial stabiliser. With $v=(-2,-1,0,1,2)$, the inputs and states have integer entries in $[-4,4]$, the readout rows entries in $[-2,2]$, and the biases and logits are integers in $[-10,10]$, so the construction runs exactly in 8-bit integer arithmetic with 32-bit accumulation at every length.

\subsection{Fixed vectors of a contraction}
\label{app:proof-fixed}

\begin{lemma}
\label{lem:fixed}
If $\|A\|_2\le1$ and $Aw=w$, then $A^{\T}w=w$.
\end{lemma}

\begin{proof}
$\|A^{\T}w-w\|^2=\|A^{\T}w\|^2-2\langle A^{\T}w,w\rangle+\|w\|^2=\|A^{\T}w\|^2-2\langle w,Aw\rangle+\|w\|^2=\|A^{\T}w\|^2-\|w\|^2\le0$, since $\|A^{\T}\|_2=\|A\|_2\le1$.
\end{proof}

\subsection{Reset limit of the token layer}
\label{app:proof-noreset}

\begin{theorem}[Obstruction for token-only non-expanding affine layers]
\label{thm:noreset}
Let the token layer have any experts with $\|A_x\|_2\le1$, any output that depends only on $(\h{0}_t,x_t)$, and finitely many reachable states. Suppose a task has words $\gamma$, $\alpha$ and $\chi$ such that for all $u\in\gamma\alpha\{\alpha,\chi\}^*$ and $u'\in\gamma\chi\{\alpha,\chi\}^*$ some nonempty word $w$ gives $uw$ and $u'w$ different labels. Then the token layer does not track the task.
\end{theorem}

\begin{proof}[Proof of \cref{thm:noreset}]
Suppose the token layer tracks the task. The words $\alpha$ and $\chi$ are nonempty, since for empty $\alpha$ the words $u=u'=\gamma\chi$ would be equal.

\emph{Step 1 (two disjoint invariant sets).} Let $Z_\alpha=\{h(\gamma\alpha y):y\in\{\alpha,\chi\}^*\}$ and $Z_\chi=\{h(\gamma\chi y):y\in\{\alpha,\chi\}^*\}$, both finite. They are disjoint, since $h(u)=h(u')$ would give $uw$ and $u'w$ the same state $F_w(h(u))$ and the same last token, hence the same output, while their labels differ. With $o=h(\gamma)$, $F_\alpha$ and $F_\chi$ preserve $Z_\alpha$ and $Z_\chi$ and send $o$ into $Z_\alpha$ and $Z_\chi$ respectively, so they map $Z=Z_\alpha\cup Z_\chi\cup\{o\}$ into itself.

\emph{Step 2 (an idempotent).} Every self-map $s$ of the finite set $Z$ has an idempotent power $s^j$, $j\ge1$, since its powers eventually cycle and the cycle contains $s^j$ for a multiple $j$ of its length that is at least the index. Let $e_\alpha=(F_\alpha|_Z)^{j_\alpha}$, $e_\chi=(F_\chi|_Z)^{j_\chi}$ and $e=(e_\chi\circ e_\alpha)^{j}$ be idempotent. The leftmost factor of $e$ is $e_\chi=e_\chi\circ e_\chi$, so
\[
e_\chi\circ e=e,\qquad\text{hence}\qquad e\circ e_\chi\circ e=e\circ e=e\quad\text{on }Z.
\]

\emph{Step 3 (two maps with common fixed points).} The word $y_e=(\alpha^{j_\alpha}\chi^{j_\chi})^{j}$, which starts with $\alpha$, applies $e$. The affine maps $\Phi=F_{y_e}$ and $\Psi=F_{y_e}\circ F_\chi^{j_\chi}$ have linear parts $A_\Phi$ and $A_\Psi$ of norm at most $1$, and $\Phi=e$ and $\Psi=e\circ e_\chi$ on $Z$. Then $p=\Phi(o)=h(\gamma y_e)\in Z_\alpha$ and $q=\Psi(o)=h(\gamma\chi^{j_\chi}y_e)\in Z_\chi$, so $p\ne q$, and by Step 2
\[
\Phi(p)=e(e(o))=p,\quad\Phi(q)=e(e(e_\chi(o)))=q,\quad\Psi(q)=(e\circ e_\chi\circ e)(e_\chi(o))=q,\quad\Psi(p)=(e\circ e_\chi\circ e)(o)=p.
\]

\emph{Step 4 (contradiction).} From $\Phi(o)=\Phi(p)=p$ and $\Phi(q)=q$, $A_\Phi(p-o)=0$ and $A_\Phi(q-p)=q-p$, so $A_\Phi^{\T}(q-p)=q-p$ by \cref{lem:fixed} and $\langle p-o,q-p\rangle=\langle A_\Phi(p-o),q-p\rangle=0$. Likewise $\Psi(o)=\Psi(q)=q$ and $\Psi(p)=p$ give $\langle q-o,p-q\rangle=0$. Adding gives $\langle (p-o)-(q-o),\,q-p\rangle=-\|q-p\|^2=0$, which contradicts $p\ne q$.
\end{proof}

\emph{Connection to \cref{thm:contextual-memory}.} The first-symbol impossibility follows from the example below. The positive construction in \cref{prop:evenpairs} stores the first symbol in $\xi_t$; reading $\xi_t$ directly gives the first-symbol label, and the stated token-dependent readout gives even pairs.

\emph{Examples.} ``First symbol'' (label $x_1$) and even pairs over $\{\mathtt a,\mathtt b\}$ satisfy the hypothesis with $\gamma$ empty, $\alpha=\mathtt a$, $\chi=\mathtt b$ and $w=\mathtt a$, which gives the labels $\mathtt a$ and $\mathtt b$, or $0$ and $1$. Modular arithmetic satisfies it with $\gamma=1$, $\alpha={+}0$, $\chi={\times}0$ and $w={+}0$, since after $u$ the finished terms sum to $1$ and the current term is $0$, after $u'$ both are $0$, and $w$ gives the labels $1$ and $0$. No group task with the running product as label satisfies it, since if the products $g_\alpha$ and $g_\chi$ of $\alpha$ and $\chi$ have orders $o_\alpha$ and $o_\chi$, then $\gamma\alpha^{o_\alpha}$ and $\gamma\chi^{o_\chi}$ both reach the state of $g_\gamma$.

\subsection{Even pairs and modular arithmetic}
\label{app:proof-evenpairs}

\begin{proposition}[Resets with one routed layer]
\label{prop:evenpairs}
A token layer and one routed layer of router depth $D{=}2$, with $b\ge2$, track even pairs over $\{\mathtt a,\mathtt b\}$, whose label is $1$ if $x_t\ne x_1$ and $0$ otherwise. A token layer and one routed layer of router depth $D{=}7$, with $b\ge10$ and $R\ge5$, track modular arithmetic modulo $5$ with ${+}$, ${-}$, ${\times}$ and the usual precedence.
\end{proposition}

\begin{proof}[Proof of \cref{prop:evenpairs} for even pairs]
\emph{Token layer.} One expert, used by every token, acts on two coordinates $(\kappa,\zeta)$ of one block ($b\ge2$) with transition $\begin{psmallmatrix}0&0\\1&0\end{psmallmatrix}$, a partial permutation (a rotation by $90^\circ$ followed by $\delta_\kappa=0$), and input $(1,0)$. Hence $(\kappa,\zeta)\leftarrow(1,\,\text{old }\kappa)$, giving $(1,0)$ at $t=1$ and $(1,1)$ for $t\ge2$, and $\kappa$ meets the condition of \cref{lem:kappa}.

\emph{Tree router.} The top node tests $\zeta-\tfrac12\kappa>0$, which by \cref{lem:kappa} goes right exactly when $t\ge2$, with margin $\tfrac12$ before the division. Both second-level nodes test the token with $e(\mathtt a)=1$, $e(\mathtt b)=-1$ and go right when $x_t=\mathtt a$, giving the experts (first, $\mathtt a$), (first, $\mathtt b$), (later, $\mathtt a$) and (later, $\mathtt b$).

\emph{Routed layer.} On one coordinate $\xi$, first-step experts have transition $0$ and input $e(x_1)$ and later experts transition $1$ and input $0$, so $\xi_t=e(x_1)$ for $t\ge1$.

\emph{Output.} With $R\ge1$, the expert for the current token $y$ has first row of $P_r$ equal to $-e(y)\,\mathbf e_\xi^{\T}$ and other rows $0$, $Q$ has the single nonzero entry $1$ at row $1$ and column $1$, and $\beta_{\out}=0$. Class $1$ ($x_t\ne x_1$) has logit $-e(y)e(x_1)=\pm1$, positive exactly when $x_t\ne x_1$, and class $0$ has logit $0$, a margin of $1$.
\end{proof}

\begin{proof}[Proof of \cref{prop:evenpairs} for modular arithmetic]
\emph{Task.} Residues $0,\dots,4$ come at odd steps and operators ${+},{-},{\times}$ at even steps, with $\mathrm{acc}_1=0$ and $\mathrm{term}_1=x_1$. At odd $t\ge3$ with $o=x_{t-1}$ and $r=x_t$, $o\in\{{+},{-}\}$ gives $\mathrm{acc}_t=\mathrm{acc}_{t-2}+\mathrm{term}_{t-2}$ and $\mathrm{term}_t=\pm r$, and $o={\times}$ gives $\mathrm{acc}_t=\mathrm{acc}_{t-2}$ and $\mathrm{term}_t=\mathrm{term}_{t-2}\,r$, all modulo $5$. The label at odd $t$ is $\mathrm{acc}_t+\mathrm{term}_t$, and even steps are unlabelled.

\emph{Token layer.} One block ($b\ge10$) holds two areas $u^{\mathrm t}$ and $u^{\mathrm f}$ of $5$ coordinates for one-hot codes over $\Z_5$ of the current term and the pending factor, with $u_0=(0,\mathbf e_1)$ under offset coding. Let $\Pi_r$ be the permutation $\mathbf e_j\mapsto \mathbf e_{rj}$ for $r\ne0$. A residue $r\ne0$ sets $u^{\mathrm t}\leftarrow\Pi_ru^{\mathrm f}$, the residue $0$ sets $u^{\mathrm t}\leftarrow \mathbf e_0$, and both set $u^{\mathrm f}\leftarrow0$. The operators ${+}$ and ${-}$ keep $u^{\mathrm t}$ and set $u^{\mathrm f}\leftarrow \mathbf e_1$ and $\mathbf e_4$, and ${\times}$ sets $u^{\mathrm f}\leftarrow u^{\mathrm t}$ and $u^{\mathrm t}\leftarrow0$. Each is a partial permutation within the block (\cref{fact:set}). By induction, $(u^{\mathrm t},u^{\mathrm f})$ is $(\mathbf e_{\mathrm{term}_t},0)$ after an odd step, $(\mathbf e_{\mathrm{term}_{t-1}},\mathbf e_{\pm1})$ after ${\pm}$ and $(0,\mathbf e_{\mathrm{term}_{t-1}})$ after ${\times}$, since $u^{\mathrm t}=\Pi_{x_1}\mathbf e_1=\mathbf e_{x_1}$ at $t=1$ and, at odd $t\ge3$, $\Pi_r\mathbf e_{\pm1}=\mathbf e_{\pm r}$ after ${\pm}$ and $\Pi_r\mathbf e_{\mathrm{term}_{t-2}}=\mathbf e_{r\,\mathrm{term}_{t-2}}$ after ${\times}$, with $\mathbf e_0$ for $r=0$. As $u_0$ has no $u^{\mathrm t}$ component, the run state $h=u-u_0$ agrees with $u$ on $u^{\mathrm t}$.

\emph{Tree router.} Levels $1$--$4$ test the bits of the token code. Our implementation has $V=9$ with one token unused, and with $V=8$ three levels and $D=6$ suffice. Levels $5$--$7$ find the index $j$ of the one-hot $u^{\mathrm t}$, the node for bit $\iota$ testing $\sum_{j'\in S}u^{\mathrm t}_{j'}-\sum_{j'\notin S}u^{\mathrm t}_{j'}>0$ with $S=\{j':\text{bit }\iota\text{ of }j'\text{ is }1\}$. At residue and ${\pm}$ steps $u^{\mathrm t}$ is one-hot, the value is $\pm1$ and the division keeps its sign (\cref{lem:kappa} with $\beta=0$). At ${\times}$ steps every test has value $0$, and the route does not matter because all ${\times}$ experts are equal.

\emph{Routed layer.} It holds a $5$-coordinate accumulator with $u_0=\mathbf e_0$ (offset coding) and a constant coordinate $\kappa$. Experts with $x_t\in\{{+},{-}\}$ and index $j$ apply $\mathbf e_a\mapsto \mathbf e_{a+j}$, the others keep it, so by induction it is $\mathbf e_{\mathrm{acc}'_t}$, with $\mathrm{acc}'_t$ the sum of all terms followed by an operator ${\pm}$ at a step $\le t$, which equals $\mathrm{acc}_t$ at odd $t$.

\emph{Output.} At a residue step the expert knows $j=\mathrm{term}_t$, and with $Q=(I_5\ \ 0)\in\R^{5\times R}$ and $R\ge5$, $P_r$ shifts the accumulator by $j$ and puts the shifted offset $\mathbf e_{j}$ into the column of $\kappa$, so the label has logit $1$ and every other class $0$, a margin of $1$.
\end{proof}

\emph{Exact stores.} Both constructions store with $\delta=0$ and keep with $\delta=1$, whereas $\delta=0.1$ would keep $10\%$ of the old value at every store. The $d=0$ block of the initialisation (\cref{app:training}) provides exact stores in every expert.

\subsection{Finite automata}
\label{app:proof-automata}

\emph{Outline.} The construction assigns one routed layer to each cascade level and a final layer to the readout. Each component state has a one-hot encoding, and expert transitions implement its permutations and resets as partial permutations (\cref{fact:set}). A cascade reads lower-level states from before the current token, so the construction preserves enough information to recover those states after the lower layers have updated. \Cref{thm:automata-resources} gives the width and depth bounds.

\emph{Setting.} By the Krohn--Rhodes theorem \citep{krohn1965}, every finite automaton is a homomorphic image of the reachable part of a cascade of permutation--reset automata $\mathcal A_1,\dots,\mathcal A_J$ over the same input alphabet. Automaton $\mathcal A_i$ has states $\{1,\dots,m_i\}$, initial state $q_{i,0}$ and level $\lambda(i)\in\{1,\dots,\ell\}$. At step $t$, $x_t$ and the states $q_{j,t-1}$ \emph{before} step $t$ of all $j$ with $\lambda(j)<\lambda(i)$ decide whether $\mathcal A_i$ applies a permutation $\Pi$ or the constant map to a state $\hat q$. The image map commutes with the steps on reachable states, so the label at step $t$ is a function $Y(q_{1,t},\dots,q_{J,t})\in\{0,\dots,C-1\}$. Let $J_l$ be the number of automata in level $l$ and $J_{<l}$ the number below it.

\begin{theorem}[Finite-automaton construction]
\label{thm:automata-resources}
Let a finite automaton be the image of the reachable part of a cascade of $J$ permutation--reset automata in $\ell$ levels, automaton $i$ having $m_i\le m$ states, where $2m$ divides $b$. Then a model with $L=\ell+1$ routed layers, $n\ge2Jm+2$, $R\ge2$ and router depth $D=1+\lceil\log_2V\rceil+\sum_i\lceil\log_2m_i\rceil$ tracks the automaton.
\end{theorem}
For a cascade of $J$ automata with $m_i$ states, the router depth $D=1+\lceil\log_2V\rceil+\sum_i\lceil\log_2m_i\rceil$ gives a bank of $2^D\ge2V\prod_im_i$ experts.

\begin{proof}[Proof of \cref{thm:automata,thm:automata-resources}]
\emph{State layout.} The token layer holds $(\kappa,\zeta)$ as in \cref{app:proof-evenpairs}, with $\kappa=1$ and $\zeta=[t\ge2]$ after step $t$. For each $\mathcal A_i$ with $\lambda(i)=l\le\ell$, routed layer $l$ holds $2m$ coordinates inside one block, the first $m_i$ forming the \emph{current} area $u_i$ and coordinates $m{+}1,\dots,m{+}m_i$ the \emph{previous} area $\bar u_i$, and since $2m$ divides $b$ these tile whole blocks. For each $\lambda(j)<l$ it holds a \emph{relay} $\tilde u_j$ of $m_j$ coordinates, plus $\zeta$ and $\kappa$, in total $2mJ_l+\sum_{\lambda(j)<l}m_j+2\le2m(J_l+J_{<l})+2\le2Jm+2\le n$ coordinates. Relays, $\zeta$ and $\kappa$ have transition $0$, so their position is free. The collector, routed layer $\ell{+}1$, uses only $\kappa$.

\emph{Invariant.} For $t\ge1$, after step $t$, routed layer $l\le\ell$ holds $u_i=\mathbf e_{q_{i,t}}$, $\bar u_i=\mathbf e_{q_{i,t-1}}$ if $\mathcal A_i$ was reset at step $t$ and $\bar u_i=0$ otherwise, $\tilde u_j=\mathbf e_{q_{j,t-1}}$ for every $\lambda(j)<l$, $\zeta=[t\ge2]$ and $\kappa=1$. We induct over $t$, and within a step over $l$.

\emph{Tree router of routed layer $l$.} It reads the token and the normalised state of layer $l-1$ after step $t$, which satisfies the invariant (for $l=1$, the token layer). Level $1$ tests $\zeta-\tfrac12\kappa>0$, of value $\pm\tfrac12$, so by \cref{lem:kappa} it goes right exactly when $t\ge2$. The next $\lceil\log_2V\rceil$ levels read $x_t$. On the right, for each $\lambda(j)<l-1$, $\lceil\log_2m_j\rceil$ levels read $q_{j,t-1}$ from $\tilde u_j$ by binary search, testing ``$q\in S$'' as $\sum_{q\in S}\tilde u_{j,q}-\sum_{q\notin S}\tilde u_{j,q}>0$. For each $\lambda(j)=l-1$, the path then knows the inputs of $\mathcal A_j$ at step $t$, hence whether it was reset or permuted by $\Pi$. After a reset it reads $q_{j,t-1}$ from $\bar u_j$, and after a permutation it tests $q_{j,t}\in\Pi(S)$ on $u_j$ by the same rule, which is equivalent to $q_{j,t-1}\in S$. All these tests have no offset and value $\pm1$, so the division keeps their sign (\cref{lem:kappa} with $\beta=0$). The right path thus knows $x_t$ and all $q_{j,t-1}$ with $\lambda(j)<l$, and the left path ($t=1$) knows $x_1$. This takes $1+\lceil\log_2V\rceil+\sum_{\lambda(j)<l}\lceil\log_2m_j\rceil\le D$ levels, padded to $D$.

\emph{Experts of routed layer $l$.} On the right, the expert knows for each $\lambda(i)=l$ whether $\mathcal A_i$ permutes by $\Pi$ or resets to $\hat q$, and applies
\[
\begin{pmatrix}u_i\\\bar u_i\end{pmatrix}\leftarrow\begin{pmatrix}\Pi&0\\0&0\end{pmatrix}\begin{pmatrix}u_i\\\bar u_i\end{pmatrix}
\qquad\text{or}\qquad
\begin{pmatrix}u_i\\\bar u_i\end{pmatrix}\leftarrow\begin{pmatrix}0&0\\I&0\end{pmatrix}\begin{pmatrix}u_i\\\bar u_i\end{pmatrix}+\begin{pmatrix}\mathbf e_{\hat q}\\0\end{pmatrix},
\]
partial permutations inside one block and hence transitions (\cref{fact:set}). On $\tilde u_j$, $\zeta$ and $\kappa$ it has transition $0$ and inputs $\mathbf e_{q_{j,t-1}}$, $1$ and $1$. From the invariant after step $t-1$, a permutation gives $u_i=\Pi \mathbf e_{q_{i,t-1}}=\mathbf e_{q_{i,t}}$ and $\bar u_i=0$, and a reset gives $u_i=\mathbf e_{\hat q}=\mathbf e_{q_{i,t}}$ and $\bar u_i=\mathbf e_{q_{i,t-1}}$. On the left ($t=1$) the expert knows $q_{i,1}$ and whether $\mathcal A_i$ was reset, has transition $0$ and writes the invariant values as its input, which proves the base case.

\emph{Collector and output.} Routed layer $\ell{+}1$ routes as a layer at level $\ell{+}1$, reading levels below $\ell$ from the relays and level $\ell$ from the current and previous areas, in exactly $D$ levels. Its expert thus knows $x_t$ and the cascade state before step $t$, hence the label $y=Y(q_{\cdot,t})$. With $\vartheta(y)=(\cos\frac{2\pi y}{C},\sin\frac{2\pi y}{C},0,\dots,0)\in\R^R$, $P_r=\vartheta(y)\mathbf e_\kappa^{\T}$, row $y$ of $Q$ equal to $\vartheta(y)^{\T}$, $\beta_{\out}=0$ and all other output maps $0$, the logit of class $y'$ is $\cos(2\pi(y'-y)/C)$, maximal only at $y'=y$ with margin $1-\cos(2\pi/C)$ for $C\ge2$ (and $C=1$ is trivial).
\end{proof}

\subsection{Diagonal stacks}
\label{app:proof-diagonal}

\begin{proposition}[Diagonal stacks]
\label{prop:diagonal}
Let every layer of a stack have diagonal transitions (real or complex), with transition and input chosen from the token and the states of the previous layers, and let each layer reach finitely many states. If the tokens generate $A_5$, the stack does not track the running product in $A_5$.
\end{proposition}

The proposition is the recurrent form of a classical fact of Krohn--Rhodes theory, that cascades of abelian groups and resets realise only solvable groups \citep{krohn1965}. The proof shows that a stack of diagonal layers with finitely many reachable states is such a cascade.

Let $Z_l$ be the finite set of reachable state tuples of layers $0,\ldots,l$, $\tau_w^l$ the transformation of $Z_l$ induced by a word $w$, and $\mathcal S_l=\{\tau_w^l\}$. For $l\ge1$, $\pi_l:Z_l\to Z_{l-1}$ drops the top state, which is surjective and commutes with every $\tau_w$, so $\theta_l(\tau_w^l)=\tau_w^{l-1}$ is a semigroup homomorphism.

\begin{proof}[Proof of \cref{prop:diagonal}]
\emph{Step 1 (diagonal permutation groups are abelian).} Let $Y\subset\mathbb C^n$ be finite and $\Gamma$ a group of permutations of $Y$, each the restriction of some $h\mapsto\Sigma_\sigma h+c_\sigma$ with $\Sigma_\sigma$ diagonal. On the span $U$ of the differences of $Y$, $\Sigma_\sigma(y-y')=\sigma(y)-\sigma(y')$, so $\Lambda(\sigma)=\Sigma_\sigma|_U$ depends only on $\sigma$ and is a homomorphism into commuting matrices. If $\Lambda(\sigma)=I$, then $\sigma(y)=y+\tau$ for a fixed $\tau$, and summing over $Y$ gives $|Y|\tau=0$. So $\Lambda$ is injective and $\Gamma$ is abelian.

\emph{Step 2 (every subgroup of $\mathcal S_l$ is solvable).} Let $H\le\mathcal S_l$ have identity $e$ and $Y=e(Z_l)$. Each $g\in H$ satisfies $g=ge=eg$, so it is determined by $g|_Y$, a bijection of $Y$ with inverse $g^{-1}|_Y$. For $l=0$, each $g|_Y$ is the restriction of an affine map with diagonal linear part, so $H$ is abelian by Step 1. For $l\ge1$, $\theta_l(H)$ is solvable by induction. With $e'=\theta_l(e)$, $\pi_l(y)=e'(\pi_l(y))$ on $Y$, so each $g$ in the kernel $N$ of $\theta_l$ on $H$ maps every fibre $Y_{y'}=\{y\in Y:\pi_l(y)=y'\}$ onto itself. For $g=\tau^l_w$, the lower layers follow the same trajectory from every $y\in Y_{y'}$, so the top state of $g(y)$ is $\Sigma_wy_{\mathrm{top}}+c_w$ with $\Sigma_w$ diagonal. The top state determines $y$ in $Y_{y'}$, so $N$ acts on each fibre as a group of Step 1 and embeds in a product of abelian groups. So $H$ is an extension of the solvable $H/N$ by the abelian $N$.

\emph{Step 3 (a group image of a finite semigroup is the image of a subgroup).} Let $\theta:\mathcal S\to\mathcal G$ be a surjective homomorphism from a finite semigroup onto a group, and take an idempotent $e$ minimising $|e\mathcal Se|$ (as in \cref{app:proof-noreset}). The set $e\mathcal Se$ is closed under products with identity $e$. For $a\in e\mathcal Se$ let $f=a^j$, $j\ge1$, be idempotent. Then $f\mathcal Sf\subseteq e\mathcal Se$, so $f\mathcal Sf=e\mathcal Se$ by minimality, $e=fxf$ for some $x$, and $e=f(fxf)=fe=f$. Hence $a^j=e$, $a$ is invertible in $e\mathcal Se$, and $e\mathcal Se$ is a group with image $\theta(e)\theta(\mathcal S)\theta(e)=\mathcal G$, since the idempotent $\theta(e)$ is the identity.

\emph{Step 4 (conclusion).} Suppose the stack tracks the running product. For $z\in Z_L$ let $\psi(z)=g(u)$ for a word $u$ reaching $z$ (the empty word reaches the zero tuple, with $g=e$). If $u$ and $u'$ reach $z$, the outputs after $ux$ and $u'x$ agree, so $g_xg(u)=g_xg(u')$ and $\psi$ is well defined. Then $\psi(\tau^L_w(z))=g_w\psi(z)$, with $g_w$ the product of $w$, later tokens on the left, so $\tau^L_w\mapsto g_w$ is a homomorphism of $\mathcal S_L$ onto $A_5$. By Steps 3 and 2, $A_5$ is then the image of a solvable group, a contradiction.
\end{proof}

This is the finite-state form of the limits of diagonal recurrences \citep{merrill2024illusion,grazzi2025unlocking}, and the same proof bounds the derived length of every subgroup of $\mathcal S_L$ by $L{+}1$, so a stack of $L{+}1$ such diagonal layers tracks no running product in a token-generated group of derived length above $L{+}1$.

\subsection{Hidden-counter family}
\label{app:proof-distinct}

A token is $x_t=(a_t,\iota_t)$ with action $a_t\in\{1,\dots,K\}$ and increment $\iota_t\in\{0,1\}$, so $V=2K$, and $q_t=\sum_{s\le t}\iota_s\bmod M$. For a table $\varphi:\{1,\dots,K\}\times\Z_M\to\mathcal G$, task $\mathcal W_\varphi$ has labels $s_t=\varphi(a_t,q_t)\,s_{t-1}$ with $s_0=e$.

\begin{lemma}[Distinct tasks]
\label{lem:distinct}
If $\varphi\ne\varphi'$, then $\mathcal W_\varphi$ and $\mathcal W_{\varphi'}$ label some input of length at most $M$ differently. So the family $\{\mathcal W_\varphi\}$ holds $|\mathcal G|^{KM}$ distinct tasks.
\end{lemma}

\begin{proof}[Proof of \cref{lem:distinct}]
Pick $(a,q)$ with $\varphi(a,q)\ne\varphi'(a,q)$, and feed $q-1$ tokens $(1,1)$ and then $(a,1)$ if $q\ge1$, or $(a,0)$ if $q=0$. This input has length $\max(q,1)\le M$ and counter $q$ at its last step. Either an earlier label differs, or both states before the last step equal some $s$ and the last labels are $\varphi(a,q)s\ne\varphi'(a,q)s$.
\end{proof}

\subsection{Capacity}
\label{app:proof-capacity}

\begin{theorem}[Capacity]
\label{thm:capacity}
Fix $n=b=8$, $V=2$, $C=60$, $R\ge5$, a token layer and one routed layer, and all other sizes, and let $N(D)=\Theta(2^D)$ be the number of weights of the model of router depth $D$. For a finite set $\Lambda\subset\R$, let $\mathcal F_\Lambda(D)$ be the set of maps from inputs to label sequences (the argmax of \cref{eq:output} at every step, with ties broken by a fixed rule) that this model computes in exact arithmetic with every weight in $\Lambda$. Then $\log_2|\mathcal F_\Lambda(D)|\le N(D)\log_2|\Lambda|$ for every $\Lambda$, and for every $D$ there is a set $\Lambda_D$ with $|\Lambda_D|\le2^D+c_0$, for a constant $c_0$, such that $\log_2|\mathcal F_{\Lambda_D}(D)|\ge2^{D-1}\log_260$. The compute per token is $O(D)$.
\end{theorem}

\begin{proof}[Proof of \cref{thm:capacity}]
\emph{Upper bound and compute.} With all sizes but $D$ fixed (including $k$, $k_{\low}$ and the embedding size), the token layer, embeddings and shared maps have $N_0$ weights independent of $D$, so $N(D)=N_0+(2^D-1)(\dim f+1)+2^DN_{\mathrm{expert}}=\Theta(2^D)$, and there are at most $|\Lambda|^{N(D)}$ weight settings. By \cref{tab:cost}, a step costs $D$ dot products of length $\dim f$ plus compute independent of $D$.

\emph{Lower bound.} For $D=1$ take $M=1$, with every expert applying $\varphi(0)$ as below. For $D\ge2$ let $K=1$, $M=2^{D-1}$, $\mathcal G=A_5$ and $\theta=2\pi/M$. We show that the model tracks every $\mathcal W_\varphi$, $\varphi:\Z_M\to A_5$, with weights in one set $\Lambda_D$, which by \cref{lem:distinct} gives $|\mathcal F_{\Lambda_D}(D)|\ge60^{2^{D-1}}$.

\emph{Token layer.} Coordinates $1,2$ hold the counter and coordinate $3$ is $\kappa$. Token $1$ applies $\mathrm{Rot}_\theta$ on coordinates $1,2$ with input $\mathrm{Rot}_\theta v-v$, $v=(1,0)$, token $0$ the identity with input $0$, and both have transition $0$ and input $1$ on coordinate $3$ (transitions by \cref{fact:set}). By \cref{thm:groups} for $\Z_M$, as $\mathrm{Rot}_\theta$ has order $M$ and only $q\equiv0$ fixes $v$, $(h_1,h_2)=(\cos\theta_t-1,\sin\theta_t)$ with $\theta_t=q_t\theta$, and $h_3=1$.

\emph{Tree router.} The top node tests $(1-\cos\theta_t)-\tfrac12(1-\cos\theta)>0$, written as $-h_1-\tfrac12(1-\cos\theta)h_3>0$. Its value is $-\tfrac12(1-\cos\theta)$ for $q_t=0$ and at least $\tfrac12(1-\cos\theta)$ for $q_t\ne0$, since then $\cos\theta_t\le\cos\theta$, so by \cref{lem:kappa} it goes right exactly when $q_t\ne0$. Below the left child all levels are padding and all experts are for $q=0$. For $q_t\ne0$,
\[
(h_1,h_2)=2\sin(\theta_t/2)\,(\cos\psi_t,\sin\psi_t),\qquad\psi_t=\pi/2+\theta_t/2\in(\pi/2,3\pi/2),
\]
with $\psi_t$ strictly increasing in $q_t$. For $\psi^*\in(\pi/2,3\pi/2)$, weights $(\sin\psi^*,-\cos\psi^*)$ give the value $2\sin(\theta_t/2)\sin(\psi^*-\psi_t)$, positive exactly when $\psi_t<\psi^*$ since $|\psi^*-\psi_t|<\pi$, and $\psi^*_i=\pi/2+(i-\tfrac12)\theta/2$ tests ``$q_t<i$''. A binary search over $q\in\{1,\dots,M-1\}$ uses thresholds $i\in\{2,\dots,M-1\}$ and $\lceil\log_2(M-1)\rceil\le D-1$ levels below the right child. These tests have no offset, so the division keeps their sign, and before it their margin is at least $2\sin(\pi/M)\sin(\pi/(2M))\ge4/M^2$, because $\theta_t/2\in[\pi/M,\pi-\pi/M]$, $|\psi^*_i-\psi_t|=|i-\tfrac12-q_t|\,\pi/M\in[\pi/(2M),\pi-\pi/(2M)]$ by $|i-\tfrac12-q_t|\le M-\tfrac52$, and $\sin x\ge2x/\pi$ on $[0,\pi/2]$.

\emph{Experts.} The expert for $q$ applies on coordinates $1$--$5$ of the routed layer $\rho(\varphi(q))$, with $\rho$ the $5{\times}5$ permutation representation of $A_5$ extended by the identity on coordinates $6$--$8$ (even permutations lie in $\SO(8)$), and input $\rho(\varphi(q))w-w$ for $w=(-2,-1,0,1,2)$. The routed layer reads $q_t$ after the token-layer update, so by the induction of \cref{thm:groups}, $\h{1}_t=\rho(s_t)w-w$, and the output of \cref{thm:groups} with $R\ge5$ and one $P_r$ for every expert tracks $s_t$.

\emph{Weight values.} Every weight lies in
$\Lambda_D=\{-10,\dots,10\}\cup\{\pm\tfrac12\}\cup\{\pm\theta,\ \cos\theta-1,\ \sin\theta,\ -\tfrac12(1-\cos\theta)\}\cup\{\sin\psi^*_i,\,-\cos\psi^*_i:2\le i\le M-1\}\cup\Lambda_{A_5}$,
where $\Lambda_{A_5}$ holds the entries of fixed skew-symmetric logarithms of the $60$ matrices $\rho(g)\oplus I_3$ (\cref{fact:set}) and $\mathrm{Rot}_\theta$ is the exponential of the skew-symmetric matrix with entries $\pm\theta$. So $|\Lambda_D|\le2(M-2)+c_0'\le2^D+c_0$.
\end{proof}

With $D+O(1)$ bits per weight, the bounds give $2^{D-1}\log_260\le\log_2|\mathcal F_{\Lambda_D}(D)|\le\log_2(2^D+c_0)\,N(D)$, so in exact arithmetic the number of represented functions is exponential in the compute per token and within a factor $O(D)$ of the capacity of $N(D)$ weights at that precision.

\subsection{Separation}
\label{app:proof-separation}

\begin{theorem}[Separation]
\label{thm:separation}
Let $M=p_1\cdots p_I$ be a product of distinct primes, and let $\mathcal G$ have a faithful orthogonal representation of dimension $m$. With $b\ge\max(m,\max_ip_i)$, $n=(I{+}1)b$, $R\ge m$ and $D=\lceil\log_2K\rceil+\sum_i\lceil\log_2p_i\rceil$, a token layer and one routed layer track every $\mathcal W_\varphi$, and for fixed $\mathcal G$ and the first $I$ primes their compute per token is polynomial in $\log M$ and $\log K$. Call a model \emph{dense} if its architecture is fixed, every step reads all its $N$ weights other than its token embedding and one $d_e$-dimensional row of that embedding, and each number has $\omega$ bits. A dense model that tracks at least a fraction $2^{-c}$ of the family has compute per token at least $d_e+N\ge(M\log_2|\mathcal G|-c/K)/(2\omega)$.
\end{theorem}

\begin{proof}[Proof of \cref{thm:separation}, construction]
\emph{Token layer.} By the Chinese remainder theorem, $q_t$ is determined by $r_{i,t}=q_t\bmod p_i$. Block $i\le I$ holds the offset code $\mathbf e_{r_{i,t}}-\mathbf e_0$ in its first $p_i$ coordinates. Increment $1$ applies the cyclic shift $\Pi_i:\mathbf e_j\mapsto \mathbf e_{j+1\bmod p_i}$ with input $\Pi_i\mathbf e_0-\mathbf e_0$, and increment $0$ the identity with input $0$, which is \cref{thm:groups} for $\Z_{p_i}$ with the regular representation and $v=\mathbf e_0$. Block $I{+}1$ holds $\kappa$. Each transition is block-diagonal with permutations or partial permutations, hence an expert transition (\cref{fact:set}).

\emph{Tree router.} The first $\lceil\log_2K\rceil$ levels read the action code of $x_t$. Then, for each $i$, $\lceil\log_2p_i\rceil$ levels read the bits of $r_{i,t}$, testing bit $\iota$ as $\sum_{r\in S}h_r-\tfrac12\kappa>0$ over block $i$ with $S=\{r:\text{bit }\iota\text{ of }r\text{ is }1\}$. As $0\notin S$, the sum is $[r_{i,t}\in S]$, the value is $\pm\tfrac12$, and by \cref{lem:kappa} the division keeps the route. The expert reached knows $a_t$ and $q_t$.

\emph{Experts and output.} The expert for $(a,q)$ applies $\rho(\varphi(a,q))$ on the first $m$ coordinates of one block of the routed layer, with input $\rho(\varphi(a,q))w-w$ for a $w$ with trivial stabiliser. As in \cref{thm:groups}, $\h{1}_t=\rho(s_t)w-w$, and its readout with $R\ge m$ and one $P_r$ for every expert tracks $s_t$.

\emph{Compute.} A step costs $D$ dot products of length $\dim f=O(\log K+n)$, transitions of $nb$ per layer, and terms independent of $M$ and $K$ for fixed $\mathcal G$. For the first $I$ primes, the prime number theorem gives $\log M=\Theta(I\log I)$, $\max_ip_i=\Theta(\log M)$ and $I=O(\log M)$, so $b=O(\log M+m)$, $n=O(\log^2M+m\log M)$ and $D\le\lceil\log_2K\rceil+\log_2M+I=O(\log K+\log M)$.
\end{proof}

Any packing of the residue areas into blocks of size $b$ plus one coordinate for $\kappa$ also works, using $\sum_ip_i+1$ token-layer and $m$ routed-layer coordinates. With a permutation representation and $w=(0,1,\dots,m-1)$, every materialised transition and every input, routing and readout coefficient is an integer or $\pm\tfrac12$, and the biases $\beta_{\out,g}=\langle\rho(g)w,w\rangle$ are integers of size $O(m^3)$.

\begin{proof}[Proof of \cref{thm:separation}, lower bound]
The dense model has $2Kd_e$ embedding numbers and $N$ further weights of $\omega$ bits, hence at most $2^{\omega(2Kd_e+N)}$ settings, each tracking at most one of the $|\mathcal G|^{KM}$ tasks of \cref{lem:distinct}. Tracking a fraction $2^{-c}$ thus needs $\omega(2Kd_e+N)\ge KM\log_2|\mathcal G|-c$, and as $2Kd_e+N\le2K(d_e+N)$, $d_e+N\ge(M\log_2|\mathcal G|-c/K)/(2\omega)$. A dense step reads all $N$ weights and $d_e$ embedding numbers, so its compute is at least $d_e+N$.
\end{proof}
\section{Architecture}
\label{app:architecture}

\paragraph{Expert parameters.}
Each expert $r$ of layer $l$ stores a decay vector $d_r\in\R^n$, block matrices $S_{r,g}\in\R^{b\times b}$, a token map $U_r\in\R^{n\times k}$ and offset $c_r\in\R^n$, a readout $P^{(l)}_r\in\R^{R\times n}$ and, in a routed layer, a layer-input map $U^{\low}_r\in\R^{n\times k_{\low}}$. Each layer shares a token code $v(x)\in\R^{k}$ and a readout $Q^{(l)}\in\R^{C\times R}$ across its experts, and a routed layer also shares a router token embedding $e^{(l)}(x)\in\R^{F}$ and the compression $W^{(l)}\in\R^{k_{\low}\times n}$. The token layer has $V$ experts and a routed layer has $2^D$. In training, each expert parameter of a routed layer is the sum of \cref{eq:increments}, and the saved model stores the materialised expert bank.

\paragraph{Layer update.}
Routed layer $l$ routes on the features
\[
f^{(l)}_t=[e^{(l)}(x_t);\rms(\h{l-1}_t)],\qquad
\rms(h)=\frac{h}{\sqrt{\frac1n\sum_i h_i^2+\epsilon}},
\]
reaches expert $r=r^{(l)}_t$ by the router traversal below and updates
\begin{equation}\label{eq:branch}
\h{l}_t=A_r\h{l}_{t-1}+T_r[x_t]+U^{\low}_rW^{(l)}\rms(\h{l-1}_t),
\end{equation}
which is \cref{eq:core} with $\Gamma_r=[U_r\;\,c_r\;\,U^{\low}_r]$ and $u^{(l)}_t=[v(x_t);1;W^{(l)}\rms(\h{l-1}_t)]$. The prediction sums the readouts of all layers,
\begin{equation}\label{eq:output}
p(y_t\mid x_{\le t})=\softmax\Big(\beta_{\out}+\sum_{l=0}^{L}Q^{(l)}P^{(l)}_{r^{(l)}_t}\h{l}_t\Big),
\end{equation}
where $\beta_{\out}$ is an output bias. The token layer is the case $r^{(0)}_t=x_t$ without the last term of \cref{eq:branch}, so $\Gamma_r=[U_r\;\,c_r]$ and $u^{(0)}_t=[v(x_t);1]$.

\subsection{Token layer}\label{sec:root}
The token layer stores one expert per token, $r^{(0)}_t=x_t$, and updates
\begin{equation}\label{eq:root}
\h{0}_t=A_{x_t}\h{0}_{t-1}+T_{x_t}[x_t].
\end{equation}
It is an input-switched affine network \citep{foerster2017isan} with rotation-and-decay transitions, and a token-only tree router adds no expressivity to it (\cref{prop:roottree}).

\paragraph{Router traversal.}
With node scores $s_\eta=w_\eta^{\T}f+\beta_\eta$, the traversal of a routed layer visits the nodes and selects the expert
\[
\eta_0=0,\qquad \eta_{j+1}=2\eta_j+1+[\,s_{\eta_j}(f^{(l)}_t)>0\,],\qquad r^{(l)}_t=\eta_D-(2^D-1).
\]

\section{Training and initialisation}
\label{app:training}

\paragraph{Router top-$B$.}
In training, each node score of a routed layer receives router noise $\tilde s_\eta=s_\eta+\sigma_{\mathrm{route}}\,\xi_\eta$, with $\xi_\eta\sim\mathrm{Logistic}(0,1)$ drawn independently per visited node, step and sequence, and evaluation uses the noiseless $s_\eta$. A path that turns right at the nodes in a set $\mathcal R$ and left at the others has log-probability
\[
\log\pi=\sum_{j<D}\log\sigma\big(\epsilon_j\,\tilde s_{\eta_j}/\tau\big),\qquad \epsilon_j=+1\text{ if }\eta_j\in\mathcal R,\ \epsilon_j=-1\text{ otherwise}.
\]
The top-$B$ set grows one level at a time from the top node. Each retained path is extended to both children, and the $B$ children with the largest log-probabilities are kept, always including the child chosen by hard routing on the noisy scores in slot~1. The selection carries no gradient. The retained experts $r_1,\dots,r_B$ receive weights $w_i=\softmax_i(\log\pi_{r_i})$, and a routed-layer step uses
\[
\bar A_t=\sum_{i=1}^{B}w_iA_{r_i},\qquad \bar u_t=\sum_{i=1}^{B}w_i\big(T_{r_i}[x_t]+U^{\low}_{r_i}W^{(l)}\rms(\h{l-1}_t)\big),\qquad \bar P_t=\sum_{i=1}^{B}w_iP_{r_i},
\]
with one associative scan per layer (\cref{prop:scan}). The token layer gives weight $1$ to the expert addressed by its token.

\paragraph{Straight-through routing.}
The forward pass uses the weights $w^{\mathrm{st}}=\mathbf e_1+w-\operatorname{sg}(w)$, where $\mathbf e_1$ puts weight $1$ on slot~1 and $\operatorname{sg}$ stops the gradient. The forward value is thus the hard routing of inference, and the gradient is that of the soft top-$B$ at the backward temperature $\tau$ of \cref{tab:hparams}.

\paragraph{Initialisation.}
The decays $d_r$ are uniform on $[0.6,1]$, except in the first block of every expert, where they are $0$, so every expert starts with an exact reset block. The entries of $S_{r,g}$ are $\mathcal N(0,\sigma_g^2)$ with $\sigma_g=0.05\cdot(1/0.05)^{(g-1)/(G-1)}$, so the first block starts near the identity and the last at a larger rotation, and $\exp(\Omega_{r,g})$ is computed with the matrix exponential at every step. The token codes $v(x)$ and router token embeddings $e^{(l)}(x)$ are drawn from $\mathcal N(0,1)$, the entries of $U_r$ from $\mathcal N(0,0.25/k)$, those of $c_r$ from $\mathcal N(0,0.01)$, those of $P_r$ and $W^{(l)}$ from $\mathcal N(0,1/n)$, those of $Q^{(l)}$ from $\mathcal N(0,0.01/R)$ and the node weights from $\mathcal N(0,1/(F+n))$. The node biases, the maps $U^{\low}_r$ and the bias $\beta_{\out}$ start at zero, so $\h{l-1}_t$ enters no expert input at the start. In a routed layer the expert parameters are drawn once and placed on the top node, scaled so that every expert starts equal to this draw (\cref{sec:increments}). Every layer starts from the zero state.

\subsection{Search and training protocol}\label{app:protocol}
Every model trains with Adam \citep{kingma2015adam} with $\beta_1=0.9$ and $\epsilon=10^{-8}$, gradient clipping at global norm 1 and a cosine schedule that decays to 1\% of the peak learning rate after a linear warm-up over 10\% of the steps. Every window of text or DNA starts from the zero state. The baselines search the settings of \cref{tab:search}. \ygg{} fixes $\beta_2=0.999$, uses no weight decay and searches the settings of \cref{tab:hparams}. Selection follows \cref{sec:experiments}, and \cref{tab:budget} gives the budget. Each final run keeps its best validation checkpoint, evaluated every 5k steps on 512 windows for text and DNA and every 4k steps on the state tasks. The test splits of text and DNA are cut into windows of the training length for every model and for the ablation. Training runs in fp32 without TF32, except for the Transformer, which uses bf16 autocast with FlashAttention-2 \citep{dao2023flashattention2}, and DeltaProduct, whose chunked kernel requires bf16 autocast.

\begin{table*}[!htbp]
\centering
\caption{\textbf{Search space and selected settings of \ygg{}.} On text and DNA the size fixes the shape, with router depth $D=10$, block size $b=16$, router token dimension $F=16$, input rank and layer-input rank $k=k_{\low}=n/4$ and readout rank $R=n/8$. The state tasks use the same rank rule. Values marked $\dagger$ are searched on the state tasks only. The large enwik8 model searches the learning rate in $[0.125, 20]\times10^{-3}$. Intervals are searched log-uniformly. The routed-layer learning-rate scale multiplies the learning rate of the routed layers. Text and DNA cells list small/medium/large, and state cells list $S_5$/$A_5$/parity/cycle/modular arithmetic.}
\label{tab:hparams}
\small
\setlength{\tabcolsep}{3pt}
\begin{tabular}{@{}llccc@{}}
\toprule
\textbf{Setting} & \textbf{Searched values} & \textbf{enwik8 (sizes)} & \textbf{hg38 (sizes)} & \textbf{State tasks} \\
\midrule
Peak learning rate ($\times10^{-3}$) & $[0.5, 20]$ & 4.2/8.3/0.84 & 0.86/0.75/0.84 & 5.4/6.7/3.3/15.6/2.8 \\
Routed-layer learning-rate scale & $\{0.3, 1\}$ & 1/0.3/1 & 1/1/0.3 & 0.3/0.3/1/1/1 \\
Router noise $\sigma_{\mathrm{route}}$ & $\{0, 0.05, 0.1\}$ & 0.05/0.05/0.1 & 0.05/0.05/0.1 & 0.05/0/0.1/0/0.1 \\
Backward temperature & \na & 0.5 & 0.5 & 0.5 \\
Router top-$B$ & $\{4, 8\}$ & 8/8/8 & 8/8/8 & 8/8/4/4/8 \\
Routed layers & $\{1,2,4\}^{\dagger}$ & 4 & 4 & 4/2/2/4/4 \\
State size $n$ & $\{64,128,256\}^{\dagger}$ & 64/128/256 & 64/128/256 & 128/256/64/64/128 \\
Router depth $D$ & $\{4\}^{\dagger}$ & 10 & 10 & 4 \\
\bottomrule
\end{tabular}
\end{table*}

\begin{table*}[!htbp]
\centering
\caption{\textbf{Search and training budget.} Steps are given as steps $\times$ batch $\times$ window length. Search spaces are in \cref{tab:hparams,tab:search}.}
\label{tab:budget}
\small
\setlength{\tabcolsep}{4pt}
\begin{tabular}{@{}l>{\raggedright\arraybackslash}p{3.4cm}>{\raggedright\arraybackslash}p{3.6cm}>{\raggedright\arraybackslash}p{2.6cm}c>{\raggedright\arraybackslash}p{3.0cm}@{}}
\toprule
\textbf{Task} & \textbf{Models} & \textbf{Search} & \textbf{Final training} & \textbf{Seeds} & \textbf{Evaluation} \\
\midrule
enwik8 & \ygg{}, GRU, LSTM, Mamba-2, Mamba-2 + PKM, Transformer, Gated DeltaNet, DeltaProduct-2/4 & 8 Bayesian trials per size, $10\mathrm{k}\times64\times256$, 1 seed & $80\mathrm{k}\times64\times256$ & 3 & best validation, full test split \\
hg38 & \ygg{}, GRU, LSTM, Mamba-2, Mamba-2 + PKM, Transformer, Gated DeltaNet, DeltaProduct-2/4, HyenaDNA & 8 Bayesian trials per size, $10\mathrm{k}\times16\times1024$, 1 seed & $80\mathrm{k}\times16\times1024$ & 3 & best validation, full test split \\
State tracking & all models of \cref{tab:state} & 8 Bayesian trials per task, size searched, $10\mathrm{k}$ steps of 256 sequences, 1 seed & $100\mathrm{k}$ steps of 256 sequences & 3 & best validation, test lengths 40--256 \\
Ablation & every variant of \cref{tab:ablation-state,tab:stored} & 8 Bayesian trials per variant and task, with the trial length of the full model on that task, 1 seed & as for the full model & 3 & as for the full model \\
\bottomrule
\end{tabular}
\end{table*}

\subsection{Hierarchical residual experts}\label{app:increments}

Hierarchical residual experts follow hierarchical shrinkage in decision trees \citep{agarwal2022hs} and the fast feedforward layers of UltraFastBERT, in which every node on the path contributes to the output \citep{belcak2023ultrafastbert}. They avoid the shrinking batch of hard trees, in which an independent expert would learn only from the few tokens that reach it \citep{belcak2023fff}. They leave the function class unchanged, since a sum of skew-symmetric generators is skew-symmetric and the decays are clipped after the sum.

\section{Tasks and settings}
\label{app:setup}

\textbf{Text and DNA.} enwik8 is split into 90M, 5M and 5M bytes with $V=256$ \citep{hutter2006enwik8}. hg38 is read one nucleotide at a time with $V=5$, the four bases and the unknown base, on the interval split of Basenji and HyenaDNA \citep{kelley2018basenji,nguyen2023hyenadna} (\cref{app:dna}).

\textbf{State tracking.} A word problem is a sequence of group elements labelled at each position by their running product, and on $S_5$ and $A_5$ the final answer is the final group element. Training draws fresh sequences at every step with random lengths from 3 to 40 and averages the loss over every labelled position, as in the word problems of \citet{liu2023shortcuts} and \citet{merrill2024illusion}. Validation and test sets hold 4096 and 8192 sequences with lengths drawn uniformly from 40 to 256. Search trials are validated every 2k steps. The validation set selects the trial and the checkpoint, as in \citet{walker2025slice}, and the test set gives the score. The search also sets the size of every model (\cref{tab:hparams,tab:search}).

\textbf{State-tracking results.} The learned models generalise beyond the training lengths on the tasks that the constructions of \cref{sec:expressivity} support. A token-layer block suffices for the group tasks, and contextual routing supports modular arithmetic with precedence (\cref{tab:state}). Mamba-2 remains near chance on both $S_5$ and $A_5$. DeltaProduct-4 reaches 0.935 on $S_5$, consistent with the finding that sufficiently long Householder products can learn this task \citep{siems2025deltaproduct}.

\section{Baselines}
\label{app:baselines}

\paragraph{Implementations.}
GRU and LSTM are the PyTorch modules with a token embedding and a linear output layer. Mamba-2 is the \texttt{mamba\_ssm} language model with expansion 2, convolution width 4 and, on text and DNA, head size $P=\min(64,d/4)$ rounded down to a multiple of 8 \citep{dao2024mamba2}. The Transformer is the \texttt{transformers} \citep{wolf2020transformers} Llama model \citep{touvron2023llama} with RMSNorm pre-normalisation, rotary position embeddings, a SwiGLU MLP of width $8d/3$, no biases, 2 to 4 heads and an untied output layer. Gated DeltaNet and DeltaProduct are the \texttt{fla} models \citep{yang2024fla} with their default value expansion, output gate, short convolution and MLP ratio \citep{yang2025gdn,siems2025deltaproduct}. Gated DeltaNet and DeltaProduct use negative eigenvalues on the state tasks \citep{grazzi2025unlocking}, and DeltaProduct also on text and DNA, where it adds a forget gate. HyenaDNA is the model code of its authors with $V=5$, trained from scratch \citep{nguyen2023hyenadna}. PD-SSM is the released model code of its authors with their default dictionary size of 8, trained from scratch \citep{terzic2025pdssm}. Mamba-2 + PKM is the Mamba-2 language model with one product-key memory layer \citep{lample2019pkm}, whose keys and values bring its stored parameters to those of \ygg{} at each size, and its number of heads and selected keys is searched with its other settings.

\paragraph{Active parameters.}
Active parameters are the total minus the rows of an untied input table that the current token does not read; every baseline therefore counts its whole output layer as active. For \ygg{} they are one expert per layer, the token rows of its tables, the $D$ nodes on the path of each routed layer and the shared maps. Stored parameters count every parameter of the trained model and active parameters those that one token reads, both with the token input of each expert in its factored form $U_r$, while the file size is that of the weights the timed runtime loads, which in table mode holds the precomputed rows $T_r[x]$ for every token in place of $U_r$. \Cref{tab:enwik8-full,tab:dna-full} give both counts.

\begin{table}[!htbp]
\centering
\caption{\textbf{Search spaces of the baselines.} On text and DNA the size fixes the shape (\cref{tab:shapes}). Learning rates are searched log-uniformly. GRU (enwik8 small, hg38 small and medium), LSTM (enwik8 medium), Mamba-2 (enwik8 large, hg38 medium), Gated DeltaNet (enwik8 small) and DeltaProduct-4 (enwik8 medium) search the learning rate up to $8\times10^{-2}$. DP is DeltaProduct and GDN Gated DeltaNet.}
\label{tab:search}
\small
\setlength{\tabcolsep}{3pt}
\begin{tabular}{@{}lll@{}}
\toprule
\textbf{Setting} & \textbf{Text and DNA} & \textbf{State tasks} \\
\midrule
Learning rate & $[5\times10^{-4},2\times10^{-2}]$ & $[10^{-4},2\times10^{-2}]$ \\
Weight decay on matrices & $\{0,0.01,0.1\}$ & $\{0,10^{-6},0.01,0.1\}$ \\
$\beta_2$ & $\{0.95,0.99,0.999\}$ & $\{0.95,0.99,0.999\}$ \\
Layers & fixed & 1 to 4 \\
Width & fixed & $\{64,128,256,384,512\}$ \\
Head size, Transformer & fixed & 16 to 128 \\
Head size, GDN and DP & fixed & 16 to 128, and 12 heads of 32 at every width \\
PD-SSM layers & \na & 1, 2 or 4 at width 64 or 128, 1 or 2 at width 256 \\
Mamba-2 head size & $\min(64,d/4)$ & $\{16,32,64\}$ \\
Mamba-2 state size & 128 & $\{16,32,64\}$ \\
Trial length & $10\mathrm{k}$ steps & $10\mathrm{k}$ steps \\
\bottomrule
\end{tabular}
\end{table}

\begin{table}[!htbp]
\centering
\caption{\textbf{Shapes of the models of \cref{tab:enwik8-full,tab:dna-full}.} $L$ is the number of layers, $d$ the width and $H$ the number of heads, and for Mamba-2 $N$ is the state size and $P$ the head dimension.}
\label{tab:shapes}
\small
\setlength{\tabcolsep}{4pt}
\begin{tabular}{@{}lll@{}}
\toprule
\textbf{Model} & \textbf{enwik8} & \textbf{hg38} \\
\midrule
\multicolumn{3}{@{}l}{\textit{Small}} \\
\ygg{} & 4 routed layers, $n{=}64$, $D{=}10$ & 4 routed layers, $n{=}64$, $D{=}10$ \\
GRU & $L{=}2$, $d{=}44$ & $L{=}2$, $d{=}48$ \\
LSTM & $L{=}2$, $d{=}40$ & $L{=}2$, $d{=}40$ \\
Mamba-2 & $L{=}1$, $d{=}40$, $N{=}128$, $P{=}8$ & $L{=}1$, $d{=}48$, $N{=}128$, $P{=}8$ \\
Transformer & $L{=}2$, $d{=}32$, $H{=}2$ & $L{=}2$, $d{=}32$, $H{=}2$ \\
Gated DeltaNet & $L{=}1$, $d{=}40$, $H{=}2$ & $L{=}1$, $d{=}40$, $H{=}2$ \\
DeltaProduct-2 & $L{=}2$, $d{=}24$, $H{=}2$ & $L{=}2$, $d{=}24$, $H{=}2$ \\
DeltaProduct-4 & $L{=}1$, $d{=}32$, $H{=}2$ & $L{=}1$, $d{=}32$, $H{=}2$ \\
HyenaDNA & \na & $L{=}2$, $d{=}16$ \\
\midrule
\multicolumn{3}{@{}l}{\textit{Medium}} \\
\ygg{} & 4 routed layers, $n{=}128$, $D{=}10$ & 4 routed layers, $n{=}128$, $D{=}10$ \\
GRU & $L{=}2$, $d{=}84$ & $L{=}2$, $d{=}84$ \\
LSTM & $L{=}2$, $d{=}72$ & $L{=}2$, $d{=}76$ \\
Mamba-2 & $L{=}2$, $d{=}64$, $N{=}128$, $P{=}16$ & $L{=}2$, $d{=}64$, $N{=}128$, $P{=}16$ \\
Transformer & $L{=}2$, $d{=}60$, $H{=}2$ & $L{=}2$, $d{=}60$, $H{=}2$ \\
Gated DeltaNet & $L{=}1$, $d{=}72$, $H{=}2$ & $L{=}1$, $d{=}80$, $H{=}2$ \\
DeltaProduct-2 & $L{=}2$, $d{=}48$, $H{=}3$ & $L{=}2$, $d{=}48$, $H{=}3$ \\
DeltaProduct-4 & $L{=}1$, $d{=}56$, $H{=}4$ & $L{=}1$, $d{=}56$, $H{=}4$ \\
HyenaDNA & \na & $L{=}2$, $d{=}52$ \\
\midrule
\multicolumn{3}{@{}l}{\textit{Large}} \\
\ygg{} & 4 routed layers, $n{=}256$, $D{=}10$ & 4 routed layers, $n{=}256$, $D{=}10$ \\
GRU & $L{=}2$, $d{=}152$ & $L{=}2$, $d{=}152$ \\
LSTM & $L{=}2$, $d{=}136$ & $L{=}2$, $d{=}136$ \\
Mamba-2 & $L{=}2$, $d{=}128$, $N{=}128$, $P{=}32$ & $L{=}2$, $d{=}128$, $N{=}128$, $P{=}32$ \\
Transformer & $L{=}2$, $d{=}112$, $H{=}4$ & $L{=}2$, $d{=}108$, $H{=}3$ \\
Gated DeltaNet & $L{=}2$, $d{=}96$, $H{=}3$ & $L{=}2$, $d{=}96$, $H{=}3$ \\
DeltaProduct-2 & $L{=}2$, $d{=}88$, $H{=}3$ & $L{=}1$, $d{=}120$, $H{=}4$ \\
DeltaProduct-4 & $L{=}2$, $d{=}80$, $H{=}2$ & $L{=}2$, $d{=}80$, $H{=}2$ \\
HyenaDNA & \na & $L{=}2$, $d{=}104$ \\
\bottomrule
\end{tabular}
\end{table}

\FloatBarrier
\section{Results at every size}
\label{app:sizes}

\Cref{tab:enwik8-full,tab:dna-full} report every model at the three sizes of active parameters.

\begin{table}[!htbp]
\centering
\caption{\textbf{enwik8 at the three sizes.} Total stored parameters, active parameters read per token, file size (MB) of the weights the timed runtime loads, microseconds per token on one CPU core and on a GPU at batch 1, and float test bits per byte on the full test split in windows of 256 bytes, best of 3 seeds by validation. Timing as in \cref{sec:experiments,app:gpu}, shapes in \cref{tab:shapes}, and \na{} marks untimed entries. Bold marks the best per column within each size.}
\label{tab:enwik8-full}
\resizebox{0.85\textwidth}{!}{%
\begin{tabular}{@{}llcccccc@{}}
\toprule
& \textbf{Model} & \textbf{Total} & \textbf{Active} & \textbf{File (MB)} & \textbf{CPU $\mu$s / token} $\downarrow$ & \textbf{GPU $\mu$s / token} $\downarrow$ & \textbf{Test bits / byte} $\downarrow$ \\
&  & & & & & & {\tiny\citep{hutter2006enwik8}} \\
\midrule
\multicolumn{8}{@{}l}{\textit{Small, 35k active parameters}} \\
\multirow{2}{*}{Sequential} & GRU & 46.5k & 35.3k & 0.2 & 10.0 & 40.0 & 2.371 \\
 & LSTM & 47.0k & 36.8k & 0.2 & 11.3 & 44.4 & 2.353 \\
\cmidrule(lr){1-8}
\multirow{7}{*}{Parallel} & Transformer & 43.2k & 35.0k & 0.1 & 64.0 & 193.6 & 2.298 \\
 & Mamba-2 & 42.6k & 32.4k & 1.0 & 22.9 & \textbf{29.3} & 2.436 \\
 & Mamba-2 + PKM & 16.3M & 34.1k & \na & \na & \na & 1.898 \\
 & Gated DeltaNet & 46.9k & 36.7k & \na & \na & 101.4 & 2.338 \\
 & DeltaProduct-2 & 40.7k & 34.6k & \na & \na & 147.3 & 2.222 \\
 & DeltaProduct-4 & 45.1k & 37.0k & \na & \na & 121.2 & 2.287 \\
 & \ygg{} & 16.3M & 35.5k & 154.3 & \textbf{1.4} & 57.6 & \textbf{1.667} \\
\midrule
\multicolumn{8}{@{}l}{\textit{Medium, 105k active parameters}} \\
\multirow{2}{*}{Sequential} & GRU & 128.9k & 107.5k & 0.5 & 12.8 & 60.9 & 2.077 \\
 & LSTM & 121.2k & 102.9k & 0.5 & 13.9 & 52.8 & 2.091 \\
\cmidrule(lr){1-8}
\multirow{7}{*}{Parallel} & Transformer & 117.4k & 102.1k & 0.5 & 91.2 & 236.3 & 2.036 \\
 & Mamba-2 & 120.0k & 103.7k & 1.0 & 28.3 & \textbf{45.7} & 2.039 \\
 & Mamba-2 + PKM & 54.3M & 101.2k & \na & \na & \na & 1.714 \\
 & Gated DeltaNet & 116.8k & 98.4k & \na & \na & 132.4 & 2.111 \\
 & DeltaProduct-2 & 116.0k & 103.8k & \na & \na & 130.5 & 1.964 \\
 & DeltaProduct-4 & 121.1k & 106.9k & \na & \na & 89.5 & 2.038 \\
 & \ygg{} & 54.2M & 101.8k & 317.1 & \textbf{2.5} & 76.2 & \textbf{1.559} \\
\midrule
\multicolumn{8}{@{}l}{\textit{Large, 330k active parameters}} \\
\multirow{2}{*}{Sequential} & GRU & 357.2k & 318.4k & 1.4 & 20.2 & 87.1 & 1.885 \\
 & LSTM & 368.0k & 333.3k & 1.5 & 23.7 & 125.3 & 1.869 \\
\cmidrule(lr){1-8}
\multirow{7}{*}{Parallel} & Transformer & 373.3k & 344.7k & 0.5 & 130.0 & 224.4 & 1.796 \\
 & Mamba-2 & 335.8k & 303.2k & 1.2 & 43.0 & \textbf{43.8} & 1.896 \\
 & Mamba-2 + PKM & 195.7M & 330.7k & \na & \na & \na & 1.673 \\
 & Gated DeltaNet & 348.9k & 324.4k & \na & \na & 125.5 & 1.727 \\
 & DeltaProduct-2 & 373.6k & 351.2k & \na & \na & 179.9 & 1.724 \\
 & DeltaProduct-4 & 331.4k & 311.0k & \na & \na & 162.0 & 1.740 \\
 & \ygg{} & 195.4M & 329.5k & 681.3 & \textbf{6.7} & 80.1 & \textbf{1.540} \\
\bottomrule
\end{tabular}}
\end{table}

\begin{table}[!htbp]
\centering
\caption{\textbf{hg38 at the three sizes.} Columns as in \cref{tab:enwik8-full}, per base, on the full test split of 253.9M bases in windows of 1024 bases. Best of 3 seeds by validation. HyenaDNA is timed in offline scoring on the CPU and in streaming on the GPU (\cref{app:dna}). Bold marks the best per column within each size.}
\label{tab:dna-full}
\label{tab:dna}
\resizebox{0.85\textwidth}{!}{%
\begin{tabular}{@{}llcccccc@{}}
\toprule
& \textbf{Model} & \textbf{Total} & \textbf{Active} & \textbf{File (MB)} & \textbf{CPU $\mu$s / base} $\downarrow$ & \textbf{GPU $\mu$s / base} $\downarrow$ & \textbf{Test bits / base} $\downarrow$ \\
&  & & & & & & {\tiny\citep{kelley2018basenji,nguyen2023hyenadna}} \\
\midrule
\multicolumn{8}{@{}l}{\textit{Small, 27k active parameters}} \\
\multirow{2}{*}{Sequential} & GRU & 28.7k & 28.5k & 0.1 & 9.4 & 46.1 & 1.728 \\
 & LSTM & 26.6k & 26.5k & 0.1 & 11.1 & 48.2 & 1.726 \\
\cmidrule(lr){1-8}
\multirow{8}{*}{Parallel} & Transformer & 27.1k & 27.0k & 0.2 & 111.7 & 197.3 & 1.753 \\
 & Mamba-2 & 29.4k & 29.1k & 0.9 & 21.4 & \textbf{28.2} & 1.822 \\
 & Mamba-2 + PKM & 15.6M & 23.6k & \na & \na & \na & 1.756 \\
 & Gated DeltaNet & 26.8k & 26.7k & \na & \na & 93.0 & 1.763 \\
 & DeltaProduct-2 & 28.6k & 28.5k & \na & \na & 128.7 & 1.723 \\
 & DeltaProduct-4 & 29.1k & 28.9k & \na & \na & 76.0 & 1.752 \\
 & HyenaDNA & 26.8k & 26.7k & 0.2 & 1.7 & 270.6 & 1.763 \\
 & \ygg{} & 15.6M & 25.2k & 17.9 & \textbf{1.1} & 67.4 & \textbf{1.620} \\
\midrule
\multicolumn{8}{@{}l}{\textit{Medium, 90k active parameters}} \\
\multirow{2}{*}{Sequential} & GRU & 86.5k & 86.2k & 0.3 & 12.4 & 56.3 & 1.681 \\
 & LSTM & 94.4k & 94.1k & 0.4 & 13.5 & 65.5 & 1.678 \\
\cmidrule(lr){1-8}
\multirow{8}{*}{Parallel} & Transformer & 87.3k & 87.1k & 0.4 & 184.4 & 250.5 & 1.728 \\
 & Mamba-2 & 88.3k & 87.9k & 0.9 & 27.2 & \textbf{46.5} & 1.714 \\
 & Mamba-2 + PKM & 52.1M & 82.1k & \na & \na & \na & 1.621 \\
 & Gated DeltaNet & 97.2k & 96.9k & \na & \na & 135.6 & 1.737 \\
 & DeltaProduct-2 & 91.9k & 91.7k & \na & \na & 171.0 & 1.682 \\
 & DeltaProduct-4 & 93.0k & 92.8k & \na & \na & 117.9 & 1.720 \\
 & HyenaDNA & 92.7k & 92.4k & 0.4 & 5.0 & 317.3 & 1.694 \\
 & \ygg{} & 52.0M & 81.4k & 48.1 & \textbf{1.9} & 79.7 & \textbf{1.590} \\
\midrule
\multicolumn{8}{@{}l}{\textit{Large, 290k active parameters}} \\
\multirow{2}{*}{Sequential} & GRU & 280.6k & 280.0k & 1.1 & 18.9 & 90.3 & 1.640 \\
 & LSTM & 299.5k & 298.9k & 1.2 & 22.7 & 91.5 & 1.637 \\
\cmidrule(lr){1-8}
\multirow{8}{*}{Parallel} & Transformer & 281.6k & 281.1k & 0.4 & 276.2 & 269.7 & 1.690 \\
 & Mamba-2 & 272.3k & 271.4k & 1.1 & 41.8 & \textbf{44.7} & 1.685 \\
 & Mamba-2 + PKM & 188.1M & 285.2k & \na & \na & \na & 1.598 \\
 & Gated DeltaNet & 300.7k & 300.3k & \na & \na & 126.1 & 1.653 \\
 & DeltaProduct-2 & 290.8k & 290.3k & \na & \na & 90.5 & 1.694 \\
 & DeltaProduct-4 & 291.2k & 290.9k & \na & \na & 162.1 & 1.651 \\
 & HyenaDNA & 297.7k & 297.0k & 0.5 & 8.3 & 384.0 & 1.655 \\
 & \ygg{} & 188.0M & 289.1k & 146.5 & \textbf{5.8} & 102.1 & \textbf{1.567} \\
\bottomrule
\end{tabular}}
\end{table}

\FloatBarrier
\section{Ablations}
\label{app:ablation}\label{sec:exp-ablation}

The ablations change one part of \ygg{} at a time and keep the token layer. Passing the output gives the next layer the readout $o^{(l)}_t=P_r\h{l}_t$ of the layer below, of dimension 16, in place of its state of dimension 128, both in the router features and in the layer input. The residual stream gives the next layer the sum of the projected outputs of all layers below. Routing on the output keeps the state in the layer input and gives the router the output of the layer below. The shared transition keeps one transition per routed layer, while the experts keep their own inputs and readouts. The diagonal variant removes the rotations in every layer, and the non-negative variant clips the decays to $[0,1]$ in every layer while keeping the rotations, whose complex eigenvalues lie outside $[0,1]$.

Storing outputs sets $A_r=0$ in every routed-layer expert, so the state of a routed layer is the current expert input and carries no recurrence. Its experts then store outputs, as in memory layers and Engram \citep{lample2019pkm,cheng2026engram}. Its input-map rank is raised from 32 to 49 to match the stored bytes of the full model. Every variant has its own search of 8 trials on each task on which it runs, with the protocol of the full model (\cref{tab:budget}). On enwik8 the shared transition is matched to the full model in active parameters. Every cell is the validation-best of 3 seeds.

Removing the rotations leaves $S_5$ and $A_5$ at 0.016, against chance levels of 0.008 and 0.017 (\cref{tab:ablation-state}). Clipping the decays to $[0,1]$ alone leaves the rotations to supply eigenvalues outside $[0,1]$, and $S_5$ still reaches 0.901 and parity 0.999. Doing both, so that every eigenvalue lies in $[0,1]$, sends parity, $S_5$ and $A_5$ to chance, as \cref{prop:diagonal} and \citet{grazzi2025unlocking} predict. A shared transition or token-only routing leaves modular arithmetic at 0.290 and 0.260, near its chance level of 0.2. What flows between layers does not decide state tracking, since all three such variants solve the four group tasks and keep modular arithmetic far above chance.

On text and DNA the routed-layer recurrence and contextual routing carry most of the gain (\cref{tab:stored}). Removing the routed-layer recurrence to store outputs costs 0.244 bits per byte on enwik8 and token-only routing 0.153, whereas sharing a single transition per routed layer stays within 0.003 bits per byte on enwik8 and 0.004 on hg38. Expert-specific transitions are therefore required for conditional resets on modular arithmetic (0.290 vs.\ 0.981) while contextual routing and recurrence carry most of the language-modelling gain. The state passed between layers must be wide. Passing the 16-dimensional output costs 0.040 bits on enwik8 and 0.016 on hg38, and a wide residual stream stays within 0.005 of passing the state.

\begin{table}[!ht]
\centering
\caption{\textbf{Ablations on state tracking.} Test accuracy of the best of 3 seeds by validation. Setup and bold as in \cref{tab:state}.}
\label{tab:ablation-state}
\footnotesize
\begin{tabular}{@{}lccccc@{}}
\toprule
\textbf{Variant} & $\boldsymbol{S_5}$ & $\boldsymbol{A_5}$ & \textbf{Parity} & \textbf{Cycle} & \textbf{Mod.\ arith.} \\
\midrule
Full \ygg{} & \textbf{1.000} & \textbf{0.980} & \textbf{1.000} & \textbf{1.000} & \textbf{0.981} \\
\midrule
Pass output, not state & \textbf{1.000} & \textbf{1.000} & \textbf{1.000} & \textbf{1.000} & \textbf{0.998} \\
Residual stream & \textbf{1.000} & \textbf{1.000} & \textbf{1.000} & \textbf{1.000} & 0.800 \\
Router reads output & \textbf{1.000} & \textbf{1.000} & \textbf{1.000} & \textbf{1.000} & 0.741 \\
\midrule
Shared transition & \textbf{1.000} & \textbf{1.000} & \textbf{1.000} & \textbf{1.000} & 0.290 \\
Diagonal transitions & 0.016 & 0.016 & \textbf{1.000} & \textbf{0.918} & 0.370 \\
Non-negative decay & \textbf{0.901} & \textbf{0.959} & \textbf{0.999} & \textbf{1.000} & 0.817 \\
Diagonal, non-negative decay & 0.009 & 0.015 & 0.517 & 0.866 & 0.383 \\
Route on token only & 0.896 & \textbf{1.000} & \textbf{1.000} & \textbf{1.000} & 0.260 \\
\bottomrule
\end{tabular}
\end{table}

\begin{table}[!ht]
\centering
\caption{\textbf{Ablations on language modelling.} Test bits of the medium models, best of 3 seeds by validation.}
\label{tab:stored}
\footnotesize
\begin{tabular}{@{}lcc@{}}
\toprule
\textbf{Variant} & \textbf{enwik8} $\downarrow$ & \textbf{hg38} $\downarrow$ \\
\midrule
Full \ygg{} & \textbf{1.559} & \textbf{1.590} \\
Store outputs & 1.803 & 1.689 \\
Route on token only & 1.712 & 1.676 \\
Shared transition & 1.562 & 1.594 \\
Pass output, not state & 1.599 & 1.606 \\
Residual stream & 1.564 & 1.594 \\
Router reads output & 1.575 & 1.597 \\
\bottomrule
\end{tabular}
\end{table}

\FloatBarrier
\FloatBarrier
\section{Compute per step}
\label{sec:cost}

\Cref{tab:cost} breaks one inference step of \ygg{} into its components and gives the compute of each.

\begin{table}[!htbp]
  \centering
  \caption{\textbf{Compute of one inference step.} $\dim f=F+n$ is the size of the node input and $C$ the number of classes.}
  \label{tab:cost}
  \small
  \setlength{\tabcolsep}{3pt}
  \begin{tabular}{@{}lll@{}}
    \toprule
    \textbf{Component} & \textbf{Compute} & \textbf{Grows with experts?} \\
    \midrule
    Router traversal (routed layers) & $LD$ dot products of length $\dim f$ & only through $D$ \\
    Transition (every layer) & $(L{+}1)nb$ multiply-adds & no \\
    Token input & one table row per layer & no \\
    Layer input (routed layers) & $2Lnk_{\low}$ multiply-adds & no \\
    Output (every layer) & $(L{+}1)(nR+CR)$ multiply-adds & no \\
    \bottomrule
  \end{tabular}
\end{table}

The token addresses its token-layer expert directly. In a routed layer each node test sets the address of the next node, so selecting an expert takes $D$ dependent comparisons. A dense recurrent step costs one multiply-add per stored weight.

\FloatBarrier
\section{CPU timing}
\label{app:timing}

\paragraph{Protocol.}
Every model runs on one core of an Intel Xeon Gold 6426Y (Sapphire Rapids, 2\,MB of L2 per core, 37.5\,MB of shared L3 per socket), pinned to one thread with the whole node reserved. A timed step consumes one token and computes the full distribution over all $V$ classes at batch 1. Each measurement streams real test tokens with the state carried, skips the first 10{,}000 tokens and reports the median, over 5 rounds of 2000 steps, of the mean time per token in each round. The Transformer keeps a sliding key--value cache over its training window $T$, and HyenaDNA is timed as in \cref{app:dna}. The comparison at equal stored parameters uses random weights from the initialisation of each model. Speedups are computed from unrounded times.

\paragraph{Runtimes.}
\ygg{} runs in our 8-bit integer kernel \citep{jacob2018quantization}. A step reads the token-layer expert of the token, traverses each tree router on $\rms(\h{l-1}_t)$, reads one expert per routed layer and computes the logits, with 8-bit states, routing weights and readouts, fixed-point block transitions and AVX-512 VNNI integer dot products. The kernel stores the token input of each expert either as the precomputed rows $T_r[x]$ (table mode) or as the 8-bit factor $U_r$ from which it computes $T_r[x]$ at every step (factored mode). Both modes compute the same integer function. Each baseline is timed in the fastest runtime of \cref{tab:runtimes} that passes the accuracy check. Compiled runtimes are driven by a C++ loop, and ONNX Runtime \citep{onnxruntime} binds every input and output once to preallocated buffers so that a step is a single call. At equal stored parameters \ygg{} runs in factored mode on enwik8 and in table mode on hg38 (\cref{tab:eqp_configs}).

\begin{table}[!htbp]
\centering
\caption{\textbf{CPU runtimes of the baselines.} Runtimes and precisions tried, and the fastest that passes the accuracy check, at the active sizes of \cref{tab:enwik8-full,tab:dna-full} and at the stored sizes of \cref{tab:capacity_speed}.}
\label{tab:runtimes}
\footnotesize
\setlength{\tabcolsep}{3pt}
\begin{tabular}{@{}p{0.13\linewidth}p{0.47\linewidth}p{0.36\linewidth}@{}}
\toprule
\textbf{Model} & \textbf{Runtimes tried} & \textbf{Fastest} \\
\midrule
Transformer & llama.cpp \citep{gerganov2023llamacpp} with Q8\_0 or fp32 weights, ONNX Runtime with 8-bit or fp32 weights & ONNX Runtime at five of the six active sizes, llama.cpp Q8\_0 at equal stored parameters \\
Mamba-2 & llama.cpp with Q8\_0 or fp32 weights & llama.cpp, Q8\_0 at equal stored parameters \\
GRU, LSTM & OpenVINO 2026.4 C++ API (GRUCell and LSTMCell on oneDNN, fp32 weights, fp32 or bf16 arithmetic), 8-bit TorchScript step in libtorch, ONNX Runtime with 8-bit or fp32 weights & OpenVINO at every active size, 8-bit libtorch at equal stored parameters except ONNX Runtime for the LSTM at 5M \\
HyenaDNA & libtorch in fp32 or with dynamic 8-bit linear layers & libtorch \\
\bottomrule
\end{tabular}
\end{table}

\paragraph{Bytes read per step.}\label{app:cache}
The bytes a step reads are the 64-byte lines that the kernel addresses in that step, enumerated from its own addresses and routing on the test stream. Latency follows the protocol above, once with warm caches and once with every cache evicted by reading 256\,MB before each step (\cref{tab:cache}). Over the full test stream, every routed layer uses at least 99.2\% of its $2^{10}$ experts, with routing entropy of 8.7 to 9.4 bits out of 10 on enwik8 and 4.7 to 9.8 bits on hg38.

\begin{table}[!htbp]
\centering
\caption{\textbf{Bytes read per step.} File size of the timed model, bytes of weights read per step, and microseconds per token on one CPU core measured in the cache benchmark with warm caches and with all caches evicted before every step.}
\label{tab:cache}
\footnotesize
\begin{tabular}{@{}llcccc@{}}
\toprule
\textbf{Data} & \textbf{Size} & \textbf{File (MB)} & \textbf{Read per step (KB)} & \textbf{Warm ($\mu$s/tok)} & \textbf{Evicted ($\mu$s/tok)} \\
\midrule
enwik8 & Small & 154 & 44.2 & 1.50 & 6.93 \\
 & Medium & 317 & 101.4 & 2.51 & 9.54 \\
 & Large & 681 & 280.7 & 6.76 & 17.74 \\
\midrule
hg38 & Small & 18 & 29.9 & 1.22 & 6.12 \\
 & Medium & 48 & 78.1 & 1.91 & 7.78 \\
 & Large & 147 & 239.5 & 5.78 & 15.66 \\
\bottomrule
\end{tabular}
\end{table}

\paragraph{Accuracy check.}
Each runtime is scored in bits per token on the first 32 test windows, each from an empty state, and passes when it is within 0.02 bits per token of the float model on the same windows. All reported bits are those of the float models.

\paragraph{Streaming accuracy.}
The timed kernel also runs over the full test split, 4{,}999{,}936 bytes of enwik8 and 253{,}885{,}440 bases of hg38, with the state carried from the first token to the last, as in the timing. It is then within 0.003 to 0.010 bits per token of the float model on the same stream (\cref{tab:int8stream}). Carrying the state also lowers the bits below the reported ones, which reset the state at every window.

\begin{table}[!htbp]
\centering
\caption{\textbf{Bits per token of \ygg{} over the full test split.} Reset starts every window of 256 bytes or 1024 bases from the zero state, as in the reported results. Stream carries the state over the whole split, as in the timing. The 8-bit kernel is the timed one, in table mode.}
\label{tab:int8stream}
\small
\setlength{\tabcolsep}{4pt}
\begin{tabular}{@{}llcccc@{}}
\toprule
& & \multicolumn{2}{c}{\textbf{Reset}} & \multicolumn{2}{c}{\textbf{Stream}} \\
\cmidrule(lr){3-4}\cmidrule(lr){5-6}
\textbf{Data} & \textbf{Size} & Float & 8-bit & Float & 8-bit \\
\midrule
enwik8 & Small & 1.667 & 1.676 & 1.626 & 1.635 \\
 & Medium & 1.559 & 1.565 & 1.510 & 1.516 \\
 & Large & 1.540 & 1.545 & 1.485 & 1.490 \\
\midrule
hg38 & Small & 1.620 & 1.625 & 1.614 & 1.620 \\
 & Medium & 1.590 & 1.594 & 1.583 & 1.589 \\
 & Large & 1.567 & 1.569 & 1.560 & 1.563 \\
\bottomrule
\end{tabular}
\end{table}

\paragraph{Equal stored parameters.}\label{app:eqp}
We also compare inference latency at approximately 5M, 20M, 50M, 200M and 730M stored parameters, using randomly initialised weights. \ygg{} uses state width 128 and two configurations at each size, more layers with shallower routers or fewer layers with deeper routers. We time the slower configuration. Dense models are matched to its stored parameter count within 4\%. \Cref{tab:eqp_configs} gives every shape, parameter count and file size in bytes per parameter.

At 5M stored parameters, \ygg{} is $162\times$ to $1{,}740\times$ faster per CPU step than the dense baselines, and at 730M the range is $3{,}289\times$ to $13{,}566\times$ (\cref{tab:capacity_speed}). Across these configurations, \ygg{} accesses 56k to 172k parameters per token. The gap grows because the dense implementations read their non-embedding weights on every step, while \ygg{} reads the selected paths and experts.

\begin{table*}[!htbp]
\centering
\caption{\textbf{Speed at equal stored parameters.} CPU time of \ygg{} in microseconds per token and the speedup of \ygg{} over each dense model with the same stored parameters, at batch 1 with random weights (\cref{app:timing}). The models in each row store 5.3--5.5M, 18.4--19.2M, 54.1--54.2M, 209.6--210.7M and 730--731M on enwik8, and 4.9--5.0M, 15.9--16.4M, 51.8--52.0M, 207.1--208.5M and 727--730M on hg38. Active, CPU $\mu$s and File / Q8\_0 refer to the timed \ygg{} configuration of \cref{tab:eqp_configs}. File / Q8\_0 is the size of the \ygg{} file divided by that of the Q8\_0 file of the matched Transformer.}
\label{tab:capacity_speed}
\footnotesize
\setlength{\tabcolsep}{2.6pt}
\begin{tabular}{@{}lcccccccccccccc@{}}
\toprule
& \multicolumn{7}{c}{\textbf{enwik8} ($V=256$) {\tiny\citep{hutter2006enwik8}}} & \multicolumn{7}{c}{\textbf{hg38} ($V=5$) {\tiny\citep{kelley2018basenji,nguyen2023hyenadna}}} \\
\cmidrule(lr){2-8}\cmidrule(lr){9-15}
& \multicolumn{3}{c}{\ygg{}} & \multicolumn{4}{c}{Speedup over} & \multicolumn{3}{c}{\ygg{}} & \multicolumn{4}{c}{Speedup over} \\
\cmidrule(lr){2-4}\cmidrule(lr){5-8}\cmidrule(lr){9-11}\cmidrule(lr){12-15}
\textbf{Stored params} & Active & CPU $\mu$s & \shortstack{File /\\Q8\_0} & Transf. & Mamba-2 & GRU & LSTM & Active & CPU $\mu$s & \shortstack{File /\\Q8\_0} & Transf. & Mamba-2 & GRU & LSTM \\
\midrule
5M & 56k & 1.02 & 0.98 & 518 & 473 & 174 & 162 & 62k & 0.96 & 0.86 & 1{,}740 & 421 & 173 & 165 \\
20M & 123k & 3.15 & 1.07 & 493 & 451 & 166 & 163 & 98k & 2.14 & 0.88 & 2{,}177 & 555 & 219 & 217 \\
50M & 102k & 3.76 & 1.16 & 1{,}178 & 1{,}480 & 511 & 512 & 81k & 2.25 & 0.87 & 5{,}332 & 2{,}219 & 811 & 816 \\
200M & 103k & 5.45 & 1.16 & 3{,}311 & 3{,}895 & 1{,}805 & 1{,}812 & 83k & 3.48 & 0.87 & 9{,}375 & 5{,}780 & 2{,}823 & 2{,}831 \\
730M & 172k & 11.06 & 1.17 & 5{,}146 & 6{,}214 & 3{,}300 & 3{,}289 & 139k & 7.72 & 0.86 & 13{,}566 & 9{,}131 & 4{,}683 & 4{,}721 \\
\bottomrule
\end{tabular}
\end{table*}

\begin{table*}[t]
\centering
\caption{\textbf{Configurations at equal stored parameters} (\cref{tab:capacity_speed}). Shape, stored parameters and file size in bytes per parameter (B/p) of every model. \ygg{} has $n{=}128$ and two configurations per size. The slower one is listed first, and the dense models are matched to it and timed against it.}
\label{tab:eqp_configs}
\small
\setlength{\tabcolsep}{4pt}
\begin{tabular}{@{}llrrlrr@{}}
\toprule
& \multicolumn{3}{c}{\textbf{enwik8} ($V{=}256$)} & \multicolumn{3}{c}{\textbf{hg38} ($V{=}5$)} \\
\cmidrule(lr){2-4}\cmidrule(lr){5-7}
\textbf{Model} & \textbf{Shape} & \textbf{Params} & \textbf{B/p} & \textbf{Shape} & \textbf{Params} & \textbf{B/p} \\
\midrule
\multicolumn{7}{@{}l}{\emph{5M}} \\
\ygg{} & 2 routed layers, $D{=}7$ & 5.5M & 1.05 & 3 routed layers, $D{=}7$ & 4.9M & 0.92 \\
 & 1 routed layer, $D{=}8$ & 5.4M & 1.05 & 1 routed layer, $D{=}9$ & 6.5M & 0.92 \\
Transformer & $L{=}3$, $d{=}384$, head 64 & 5.5M & 1.07 & $L{=}4$, $d{=}320$, head 64 & 5.0M & 1.06 \\
Mamba-2 & $L{=}11$, $d{=}256$, $N{=}128$, $P{=}64$ & 5.3M & 1.25 & $L{=}7$, $d{=}320$, $N{=}128$, $P{=}64$ & 4.9M & 1.25 \\
GRU & $L{=}2$, $d{=}656$ & 5.5M & 1.10 & $L{=}2$, $d{=}640$ & 4.9M & 1.01 \\
LSTM & $L{=}2$, $d{=}568$ & 5.5M & 1.01 & $L{=}2$, $d{=}552$ & 4.9M & 1.01 \\
\midrule
\multicolumn{7}{@{}l}{\emph{20M}} \\
\ygg{} & 5 routed layers, $D{=}8$ & 18.5M & 1.19 & 5 routed layers, $D{=}8$ & 16.3M & 0.92 \\
 & 3 routed layers, $D{=}9$ & 21.7M & 1.20 & 3 routed layers, $D{=}9$ & 19.5M & 0.92 \\
Transformer & $L{=}6$, $d{=}512$, head 64 & 19.2M & 1.06 & $L{=}9$, $d{=}384$, head 64 & 15.9M & 1.06 \\
Mamba-2 & $L{=}14$, $d{=}448$, $N{=}128$, $P{=}64$ & 18.9M & 1.12 & $L{=}12$, $d{=}448$, $N{=}128$, $P{=}64$ & 16.0M & 1.13 \\
GRU & $L{=}2$, $d{=}1224$ & 18.6M & 1.05 & $L{=}2$, $d{=}1168$ & 16.4M & 1.00 \\
LSTM & $L{=}2$, $d{=}1056$ & 18.4M & 1.00 & $L{=}2$, $d{=}1008$ & 16.3M & 1.00 \\
\midrule
\multicolumn{7}{@{}l}{\emph{50M}} \\
\ygg{} & 4 routed layers, $D{=}10$ & 54.2M & 1.23 & 4 routed layers, $D{=}10$ & 52.0M & 0.92 \\
 & 1 routed layer, $D{=}12$ & 54.2M & 1.23 & 1 routed layer, $D{=}12$ & 52.0M & 0.92 \\
Transformer & $L{=}9$, $d{=}704$, head 64 & 54.1M & 1.06 & $L{=}13$, $d{=}576$, head 64 & 51.8M & 1.06 \\
Mamba-2 & $L{=}25$, $d{=}576$, $N{=}128$, $P{=}64$ & 54.2M & 1.09 & $L{=}24$, $d{=}576$, $N{=}128$, $P{=}64$ & 51.8M & 1.09 \\
GRU & $L{=}2$, $d{=}2104$ & 54.2M & 1.03 & $L{=}2$, $d{=}2080$ & 52.0M & 1.00 \\
LSTM & $L{=}2$, $d{=}1824$ & 54.2M & 1.03 & $L{=}2$, $d{=}1800$ & 51.9M & 1.00 \\
\midrule
\multicolumn{7}{@{}l}{\emph{200M}} \\
\ygg{} & 4 routed layers, $D{=}12$ & 210.2M & 1.24 & 4 routed layers, $D{=}12$ & 208.0M & 0.92 \\
 & 1 routed layer, $D{=}14$ & 210.1M & 1.24 & 1 routed layer, $D{=}14$ & 207.9M & 0.92 \\
Transformer & $L{=}19$, $d{=}960$, head 64 & 210.7M & 1.06 & $L{=}13$, $d{=}1152$, head 64 & 207.1M & 1.06 \\
Mamba-2 & $L{=}36$, $d{=}960$, $N{=}128$, $P{=}64$ & 209.9M & 1.07 & $L{=}25$, $d{=}1152$, $N{=}128$, $P{=}64$ & 207.9M & 1.07 \\
GRU & $L{=}4$, $d{=}2944$ & 209.6M & 1.01 & $L{=}4$, $d{=}2944$ & 208.1M & 1.00 \\
LSTM & $L{=}4$, $d{=}2552$ & 209.8M & 1.01 & $L{=}4$, $d{=}2552$ & 208.5M & 1.00 \\
\midrule
\multicolumn{7}{@{}l}{\emph{730M}} \\
\ygg{} & 7 routed layers, $D{=}13$ & 730.0M & 1.24 & 7 routed layers, $D{=}13$ & 727.7M & 0.92 \\
 & 4 routed layers, $D{=}14$ & 833.8M & 1.24 & 4 routed layers, $D{=}14$ & 831.6M & 0.92 \\
Transformer & $L{=}28$, $d{=}1472$, head 64 & 730.2M & 1.06 & $L{=}28$, $d{=}1472$, head 64 & 729.5M & 1.06 \\
Mamba-2 & $L{=}46$, $d{=}1600$, $N{=}128$, $P{=}64$ & 730.9M & 1.07 & $L{=}59$, $d{=}1408$, $N{=}128$, $P{=}64$ & 727.9M & 1.07 \\
GRU & $L{=}4$, $d{=}5504$ & 730.0M & 1.01 & $L{=}4$, $d{=}5504$ & 727.2M & 1.00 \\
LSTM & $L{=}4$, $d{=}4768$ & 730.1M & 1.01 & $L{=}4$, $d{=}4768$ & 727.7M & 1.00 \\
\bottomrule
\end{tabular}
\end{table*}

\FloatBarrier
\section{GPU timing}
\label{app:gpu}

Every model runs on one NVIDIA A100-PCIE-40GB with torch 2.10.0 and CUDA 12.8, each baseline in its fastest public GPU path. These are the fused cuDNN kernels for GRU and LSTM, mamba\_ssm for Mamba-2, the Hugging Face Llama decode with FlashAttention and a key--value cache for the Transformer, the fla kernels for Gated DeltaNet and DeltaProduct and the Hugging Face model for HyenaDNA, streamed as in \cref{app:dna}. \ygg{} runs one fused Triton 3.6.0 kernel that updates all layers, followed by the output layer, with the expert banks precomputed. Every model runs under CUDA graphs with one step per replay, whose outputs match eager execution to within $2\times10^{-6}$, and computes the full output distribution in fp32. Each measurement is the median of 5 rounds of 512 steps after 64 warm-up steps.

\begin{figure*}[!htbp]
\centering
\includegraphics[width=\textwidth]{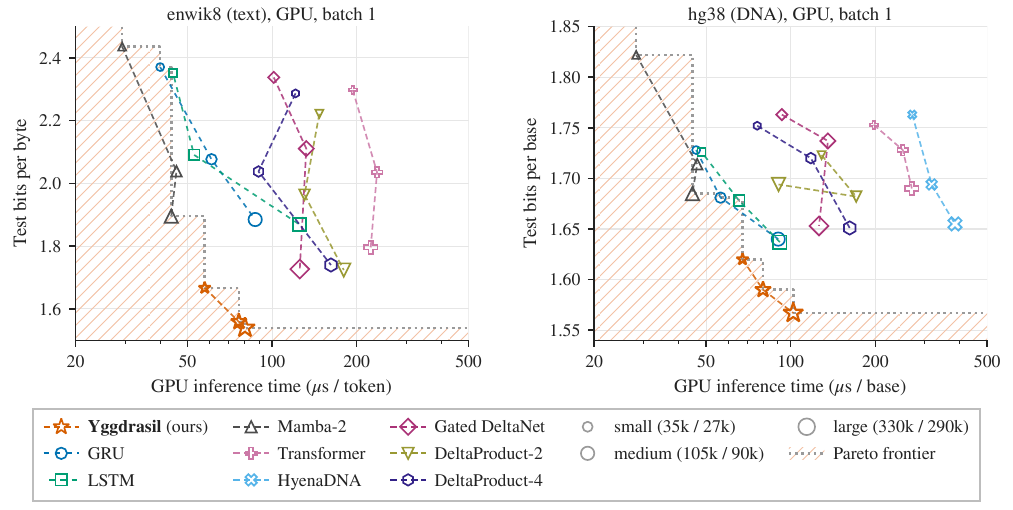}
\caption{\textbf{Accuracy against inference time on a GPU at batch 1.} Float test bits per byte on enwik8 (left) and per base on hg38 (right) against microseconds per token on one A100, from \cref{tab:gpu}. Lines and markers as in \cref{fig:pareto}. HyenaDNA is shown at its streaming time per base. The dotted staircase is the Pareto frontier, and no model reaches the hatched region.}
\label{fig:pareto-gpu}
\end{figure*}

\begin{table}[!htbp]
\centering
\caption{\textbf{GPU inference.} Microseconds per token at batch 1 and tokens per second at batch 1024, on one A100 under CUDA graphs, at the small, medium and large sizes. Bold marks the best per column within each dataset.}
\label{tab:gpu}
\resizebox{0.85\textwidth}{!}{%
\begin{tabular}{@{}llcccccc@{}}
\toprule
& & \multicolumn{3}{c}{\textbf{Batch 1, $\mu$s / token} $\downarrow$} & \multicolumn{3}{c}{\textbf{Batch 1024, tokens / s} $\uparrow$} \\
\cmidrule(lr){3-5}\cmidrule(lr){6-8}
& \textbf{Model} & Small & Medium & Large & Small & Medium & Large \\
\midrule
\multicolumn{8}{@{}l}{\emph{enwik8, 35k, 105k and 330k active parameters} {\tiny\citep{hutter2006enwik8}}} \\
\multirow{2}{*}{Sequential} & GRU & 40.0 & 60.9 & 87.1 & 12.5M & 5.84M & 1.65M \\
 & LSTM & 44.4 & 52.8 & 125.3 & 11.0M & 5.83M & 889k \\
\cmidrule(lr){1-8}
\multirow{6}{*}{Parallel} & Transformer & 193.6 & 236.3 & 224.4 & 1.39M & 685k & 398k \\
 & Mamba-2 & \textbf{29.3} & \textbf{45.7} & \textbf{43.8} & 7.77M & 3.18M & 1.77M \\
 & Gated DeltaNet & 101.4 & 132.4 & 125.5 & 9.13M & 5.29M & 2.43M \\
 & DeltaProduct-2 & 147.3 & 130.5 & 179.9 & 3.53M & 3.39M & 1.98M \\
 & DeltaProduct-4 & 121.2 & 89.5 & 162.0 & 5.03M & 4.16M & 1.81M \\
 & \ygg{} & 57.6 & 76.2 & 80.1 & \textbf{15.7M} & \textbf{7.15M} & \textbf{2.59M} \\
\midrule
\multicolumn{8}{@{}l}{\emph{hg38, 27k, 90k and 290k active parameters} {\tiny\citep{kelley2018basenji,nguyen2023hyenadna}}} \\
\multirow{2}{*}{Sequential} & GRU & 46.1 & 56.3 & 90.3 & 10.1M & 6.04M & 1.67M \\
 & LSTM & 48.2 & 65.5 & 91.5 & 11.9M & 5.51M & 895k \\
\cmidrule(lr){1-8}
\multirow{7}{*}{Parallel} & Transformer & 197.3 & 250.5 & 269.7 & 508k & 203k & 109k \\
 & Mamba-2 & \textbf{28.2} & \textbf{46.5} & \textbf{44.7} & 7.32M & 3.17M & 1.78M \\
 & Gated DeltaNet & 93.0 & 135.6 & 126.1 & 9.26M & 5.29M & 2.47M \\
 & DeltaProduct-2 & 128.7 & 171.0 & 90.5 & 3.55M & 3.42M & 2.59M \\
 & DeltaProduct-4 & 76.0 & 117.9 & 162.1 & 5.11M & 4.23M & 1.83M \\
 & HyenaDNA & 270.6 & 317.3 & 384.0 & 30.2k & 15.1k & 8.70k \\
 & \ygg{} & 67.4 & 79.7 & 102.1 & \textbf{21.8M} & \textbf{8.19M} & \textbf{2.82M} \\
\bottomrule
\end{tabular}}
\end{table}

\FloatBarrier
\section{Genomics}
\label{app:dna}

\paragraph{Split.}
We use the hg38 intervals of the Basenji BED file \citep{kelley2018basenji}, which the HyenaDNA code loads for its training, validation and test sets \citep{nguyen2023hyenadna}. No test interval overlaps a training or validation interval. HyenaDNA is trained from scratch at the three sizes with the protocol of \cref{app:protocol}.

\paragraph{HyenaDNA timing.}
The official HyenaDNA code has no recurrent decode. Its CPU time is that of offline scoring, in which one forward pass over a window of 1024 bases is divided by the window length. This is a lower bound on its time in streaming. A streaming step recomputes the window for every new base and takes 1{,}605, 5{,}003 and 7{,}297\,$\mu$s on one core at the three sizes, 876 to 997 times its offline time. Its GPU time is that of streaming.

\FloatBarrier
\section{Genomic Benchmarks}
\label{app:gb}

We follow the protocol of Caduceus \citep{schiff2024caduceus} on the eight tasks of the Genomic Benchmarks \citep{gresova2023genomic}. \ygg{} is first pretrained on the hg38 training split by next-base prediction for 10{,}000 steps at batch 1024 with windows of 1024 bases, about 10.5B bases, with reverse-complement augmentation. The model has state size 336, 4 routed layers (5 layers in total), router depth $D=10$ and input rank 84, and its learning rate of $6.23\cdot10^{-3}$ comes from the hyperparameter search. It stores 307.3M parameters and reads about 470k active parameters per base, and it reaches 1.575 test bits per base on hg38. Every fine-tune starts from this checkpoint. Each task is run on 5 cross-validation splits with seeds 1 to 5 and a 90/10 split of the training set into training and validation, with early stopping on validation accuracy. \ygg{} trains for 10 epochs at batch 256. For each task we take the learning rate in $\{10^{-3},2\cdot10^{-3}\}$ and the choice of reverse-complement augmentation with the best mean validation accuracy over the 5 splits, and report its mean test accuracy.

\Cref{tab:gb-full} adds the published bidirectional models at matched active parameter counts, Caduceus-Ph and Caduceus-PS \citep{schiff2024caduceus}, ConvNova \citep{bo2025convnova}, JanusDNA \citep{duan2025janusdna} and DNA-xLSTM-PH and DNA-xLSTM-PS \citep{schmidinger2024bioxlstm}. They read the whole sequence, or its reverse complement, before they classify it, so none of them can score a stream.

\begin{table}[!htbp]
\centering
\caption{\textbf{Genomic Benchmarks, all models.} Test accuracy (\%) under the protocol of Caduceus, mean over 5 cross-validation splits with the range over the splits after $\pm$. Params are active parameters for \ygg{} (307.3M stored) and JanusDNA. Bold marks the best per column.}
\label{tab:gb-full}
\small
\setlength{\tabcolsep}{4pt}
\resizebox{\textwidth}{!}{%
\begin{tabular}{@{}llcccccccccc@{}}
\toprule
& \textbf{Model} & \textbf{Params} & \textbf{Mouse} & \textbf{Coding} & \textbf{Hum./worm} & \textbf{Cohn} & \textbf{Ensembl} & \textbf{Reg.} & \textbf{OCR} & \textbf{Non-TATA} & \textbf{Mean} \\
& & & \multicolumn{9}{c}{\tiny\citep{gresova2023genomic}} \\
\midrule
& CNN {\tiny\citep{schiff2024caduceus}} & 264k & 71.5{\tiny$\pm$8.7} & 89.2{\tiny$\pm$0.8} & 94.2{\tiny$\pm$0.2} & 70.2{\tiny$\pm$2.1} & 74.4{\tiny$\pm$12.2} & 87.2{\tiny$\pm$0.5} & 69.8{\tiny$\pm$1.3} & 86.1{\tiny$\pm$0.9} & 80.3 \\
\midrule
\multirow{3}{*}{Causal} & HyenaDNA {\tiny\citep{schiff2024caduceus}} & 436k & 78.0{\tiny$\pm$2.5} & 90.4{\tiny$\pm$0.5} & 96.4{\tiny$\pm$0.2} & 72.9{\tiny$\pm$1.4} & 84.9{\tiny$\pm$0.6} & 86.9{\tiny$\pm$1.2} & 78.3{\tiny$\pm$0.7} & 94.4{\tiny$\pm$0.2} & 85.3 \\
 & Mamba {\tiny\citep{schiff2024caduceus}} & 468k & 74.3{\tiny$\pm$5.4} & 90.4{\tiny$\pm$0.4} & 96.7{\tiny$\pm$0.2} & 73.2{\tiny$\pm$2.9} & 86.2{\tiny$\pm$0.8} & 81.4{\tiny$\pm$21.1} & 81.5{\tiny$\pm$0.2} & 93.3{\tiny$\pm$0.7} & 84.6 \\
 & \ygg{} & 470k & 74.0{\tiny$\pm$4.1} & 93.4{\tiny$\pm$0.6} & 96.6{\tiny$\pm$0.2} & 72.9{\tiny$\pm$1.6} & 88.2{\tiny$\pm$0.6} & 87.1{\tiny$\pm$1.3} & 79.6{\tiny$\pm$0.5} & \textbf{96.0}{\tiny$\pm$1.0} & 86.0 \\
\midrule
\multirow{6}{*}{Bidirectional} & Caduceus-Ph {\tiny\citep{schiff2024caduceus}} & 470k & 75.4{\tiny$\pm$7.4} & 91.5{\tiny$\pm$0.3} & \textbf{97.3}{\tiny$\pm$0.1} & \textbf{74.7}{\tiny$\pm$0.4} & 89.3{\tiny$\pm$0.8} & 87.2{\tiny$\pm$1.1} & \textbf{82.8}{\tiny$\pm$0.6} & 94.6{\tiny$\pm$0.7} & 86.6 \\
 & Caduceus-PS {\tiny\citep{schiff2024caduceus}} & 470k & \textbf{79.3}{\tiny$\pm$5.8} & 91.0{\tiny$\pm$0.3} & 96.8{\tiny$\pm$0.2} & 74.5{\tiny$\pm$0.7} & 90.0{\tiny$\pm$0.6} & 87.3{\tiny$\pm$0.7} & 81.8{\tiny$\pm$0.6} & 94.5{\tiny$\pm$1.0} & 86.9 \\
 & ConvNova {\tiny\citep{bo2025convnova}} & 386k & 78.4{\tiny$\pm$0.9} & \textbf{94.3}{\tiny$\pm$0.1} & 96.7{\tiny$\pm$0.2} & 74.3{\tiny$\pm$0.5} & 90.0{\tiny$\pm$0.4} & 87.3{\tiny$\pm$0.2} & 79.3{\tiny$\pm$0.4} & 95.1{\tiny$\pm$0.3} & 86.9 \\
 & JanusDNA {\tiny\citep{duan2025janusdna}} & 426k & 77.0{\tiny$\pm$4.8} & 91.2{\tiny$\pm$0.3} & 97.1{\tiny$\pm$0.1} & 74.1{\tiny$\pm$0.5} & 89.7{\tiny$\pm$0.4} & \textbf{87.7}{\tiny$\pm$0.5} & 82.2{\tiny$\pm$0.3} & 95.7{\tiny$\pm$0.4} & 86.8 \\
 & DNA-xLSTM-PH {\tiny\citep{schmidinger2024bioxlstm}} & 500k & 78.0{\tiny$\pm$1.8} & 93.1{\tiny$\pm$0.1} & 96.5{\tiny$\pm$0.1} & 74.2{\tiny$\pm$0.5} & \textbf{92.0}{\tiny$\pm$0.1} & 87.2{\tiny$\pm$0.2} & 82.6{\tiny$\pm$0.2} & 95.1{\tiny$\pm$0.4} & \textbf{87.3} \\
 & DNA-xLSTM-PS {\tiny\citep{schmidinger2024bioxlstm}} & 500k & 77.8{\tiny$\pm$0.7} & 93.4{\tiny$\pm$0.2} & 95.6{\tiny$\pm$0.1} & 73.4{\tiny$\pm$0.5} & 90.2{\tiny$\pm$0.4} & 86.9{\tiny$\pm$0.5} & 80.0{\tiny$\pm$0.2} & 94.9{\tiny$\pm$0.1} & 86.5 \\
\bottomrule
\end{tabular}}
\end{table}

\end{document}